\documentclass[sn-basic]{sn-jnl}

\usepackage{subfigure}
\usepackage{adjustbox}
\usepackage{booktabs}
\usepackage{xcolor}
\usepackage{soul}
\sethlcolor{lime} 

\usepackage{graphicx}%
\usepackage{multirow}%
\usepackage{amsmath,amssymb,amsfonts}%
\usepackage{amsthm}%
\usepackage{mathrsfs}%
\usepackage[title]{appendix}%
\usepackage{xcolor}%
\usepackage{textcomp}%
\usepackage{manyfoot}%
\usepackage{booktabs}%
\usepackage{algorithm}%
\usepackage{algorithmicx}%
\usepackage{algpseudocode}%
\usepackage{listings}%

\theoremstyle{thmstyleone}%
\theoremstyle{thmstyletwo}%

\theoremstyle{thmstylethree}%

\begin{document}

\title[Article Title]{Adaptive Transformer Modelling of Density Function for Nonparametric Survival Analysis}


\author[1,2]{\fnm{Xin} \sur{Zhang}}\email{xin.zhang4@monash.edu}

\author[2,3]{\fnm{Deval} \sur{Mehta}}\email{deval.mehta@monash.edu}

\author[4]{\fnm{Yanan} \sur{Hu}}\email{Yanan.Hu@monash.edu}

\author[5]{\fnm{Chao} \sur{Zhu}}\email{chao.zhu@monash.edu}

\author[5]{\fnm{David} \sur{Darby}}\email{david.darby@monash.edu}

\author[2,3]{\fnm{Zhen} \sur{Yu}}\email{zhen.yu@monash.edu}

\author[5]{\fnm{Daniel} \sur{Merlo}}\email{daniel.Merlo@monash.edu}

\author[5]{\fnm{Melissa} \sur{Gresle}}\email{melissa.Gresle@monash.edu}

\author[5,6]{\fnm{Anneke} \sur{van der Walt}}\email{anneke.VanDerWalt@monash.edu}

\author[5,6]{\fnm{Helmut} \sur{Butzkueven}}\email{helmut.butzkueven@monash.edu}

\author*[2,3]{\fnm{Zongyuan} \sur{Ge}}\email{zongyuan.Ge@monash.edu}

\affil[1]{\orgdiv{Department of Electrical and Computer Systems Engineering}, \orgname{Monash University}, \orgaddress{\city{Melbourne}, \country{Australia}}}

\affil*[2]{\orgdiv{AIM for Health Lab}, \orgname{Monash University}, \orgaddress{\city{Melbourne}, \country{Australia}}}

\affil[3]{\orgdiv{Department of Data Science and AI}, \orgname{Monash University}, \orgaddress{\city{Melbourne}, \country{Australia}}}

\affil[4]{\orgdiv{Monash Centre for Health Research and Implementation}, \orgname{Monash University}, \orgaddress{\city{Melbourne}, \country{Australia}}}

\affil[5]{\orgdiv{Department of Neuroscience}, \orgname{Monash University}, \orgaddress{\city{Melbourne}, \country{Australia}}}

\affil[6]{\orgdiv{Department of Neurology}, \orgname{Monash University}, \orgaddress{\city{Melbourne}, \country{Australia}}}


\abstract{Survival analysis holds a crucial role across diverse disciplines, such as economics, engineering and healthcare. It empowers researchers to analyze both time-invariant and time-varying data, encompassing phenomena like customer churn, material degradation and various medical outcomes. Given the complexity and heterogeneity of such data, recent endeavors have demonstrated successful integration of deep learning methodologies to address limitations in conventional statistical approaches. However, current methods typically involve cluttered probability distribution function (PDF), have lower sensitivity in censoring prediction, only model static datasets, or only rely on recurrent neural networks for dynamic modelling. In this paper, we propose a novel survival regression method capable of producing high-quality unimodal PDFs without any prior distribution assumption, by optimizing novel Margin-Mean-Variance loss and leveraging the flexibility of Transformer to handle both temporal and non-temporal data, coined \textbf{UniSurv}. Extensive experiments on several datasets demonstrate that UniSurv places a significantly higher emphasis on censoring compared to other methods.}




\maketitle

\section{Introduction}\label{sec1}

The primary task of survival analysis is to determine the timing of one or multiple events, which can signify the moment of a mechanical system malfunction, the period of transition from corporate deficit to surplus, the instance of patient fatality or so on, depending on the specific circumstance \citep{lee2006threshold}. Among all scenarios, survival analysis for medical data poses the most severe challenges \citep{collett2023modelling}. Some medical datasets are longitudinal, as exemplified by electronic health records (EHRs), where multiple observations of each patient's covariates over time are recorded. Survival models must be capable of handling such measurements and learning from their continuous temporal trends. Moreover, observations in longitudinal data are often sparse, necessitating the effective handling of missing values for any reliable survival model, even when the missing rates are exceedingly high \citep{singer1991modeling}. 
Additionally, censoring represents a fundamental aspect of survival data, referring to cases in which complete information regarding the survival time or event occurrence of a subject is not fully observed or available within the study period \citep{leung1997censoring}. The occurrence of censoring signifies the unknown exact timing of the event, consequently lacking ground truth for comparative learning. This, in turn, poses significant challenges for deep survival learning. Existing deep learning approaches aim at mitigating this issue by typically guaranteeing non-occurrence of events before censoring. Notwithstanding, detailed elucidation pertaining to the temporal aspect of events subsequent to censoring frequently remains inadequately explored.

Developing survival analysis models requires regressing the probability of survival over a defined period. A high-quality estimation of probability distribution is essential for the time-to-event prediction.
As the initial category, parametric survival models are capable of generating high-quality probability density function (PDF) or survival curve by predetermining stochastic distribution, however, their precision is contingent upon the validity of all underlying assumptions. In contrast, non-parametric models do not presume any prior distribution of events, but they struggle to accurately predict PDF over extended temporal spans within medical datasets, consequently yielding PDF or survival curve of comparatively lower quality. 

To address the challenges in the development of survival models and to mitigate the limitations inherent in existing models, we propose \textbf{UniSurv}, a non-parametric model based on the Transformer architecture. In particular, UniSurv can: 1) generate higher quality PDF resembling normal distribution without any prior probability assumption, and significantly improve accuracy for predicting censoring by integrating novel Margin-Mean-Variance loss; 2) use distinct embedding branches for static and dynamic feature extractions separately; 3) effectively handle cases with high missing rates of longitudinal data and various data modalities. The superiority of the UniSurv is substantiated through empirical evidence obtained from real and synthetic datasets.

\section{Literature}\label{sec2}

Semi- and fully-parametric models heavily rely on the premise of making explicit assumptions about the underlying distribution of event times. They provide a structured framework for understanding the relationship between covariates and the occurrence of events over time. However, the strength of these assumptions results in overly simplistic probability distributions predicted by the models. The lack of flexibility stemming from this oversimplification also renders these models impractical in various scenarios. Cox proportional hazard (CPH) \citep{cox1972regression} is a prime example in this field. It estimates the hazard function $\lambda(t|X)$ by multiplying a predetermined base hazard function $\lambda_{0}(t)$ with the learnt representation of features $g(X)$. Subsequent studies \citep{faraggi1995neural,vinzamuri2013cox,luck2017deep} have used more sophisticated models to improve the CPH model. However, the oversimplified stochastic process continues to constrain their predictive capabilities, and it is unable to conduct dynamic analysis. Meanwhile, \cite{nagpal2021deep} introduced Deep Survival Machines (DSM), which postulates that the survival function is a composition of multiple Weibull and log-normal distributions. The parameters of those distributions are estimated by a multi-layer perceptron (MLP). Besides, \cite{nagpal2021deep2} illustrated Recurrent DSM (RDSM) by incorporating recurrent neural network (RNN) into DSM, thereby endowing it to process dynamic analysis. Nonetheless, DSM models exhibit suboptimal accuracy in predicting event times. Its loss function frequently becomes divergent during training, contributing to the overfitting problem.

Some recent works have concentrated on static analysis. For example, DeepSurv \citep{katzman2018deepsurv} employs an MLP network to replace the parametric assumptions of the hazard function present in the conventional CPH. This transformation results in a semi-parametric variant of the CPH model. The incorporation of neural networks enhances its flexibility by enabling the model to learn nonlinear relationships more adeptly from covariates. Besides, Deep Cox Mixtures (DCM) \citep{nagpal2021deep3} encounters the same underlying assumption of proportional hazards, wherein it assumes the presence of latent groups. Employing Variational Autoencoders for clustering, DCM assumes the validity of proportional hazards within each latent group. Moreover, \cite{ishwaran2008random} propose an extension of random forest (RF) algorithm, named Random Survival Forest (RSF), which initially breaks through the inherent assumptions of CPH. It computes the risk scores through the generation of Nelson-Aalen estimators within the partitions established by RF. RSF assumes independence among trees in forest, which might not always hold. This assumption can affect its performance, usually when correlations or dependencies exist among survival trees.

Several studies have explored the dynamic analysis field. \cite{lee2018deephit} propose DeepHit for competing risk events as a non-parametric model. The encoder of DeepHit is constructed as a joint MLP, while its decoder employs a series of distinct MLPs to address individual events. This design results in the generation of separate PDFs for each event. \cite{lee2019dynamic} further extend it into Dynamic DeepHit (DDH) by replacing the encoder with RNNs followed by an attention mechanism, to process longitudinal data. A primary limitation lies in the arbitrary fluctuations between adjacent predictions within the output layer, resulting in noise present in the final PDFs. This phenomenon becomes particularly pronounced when forecasting over a long-time horizon. Survival SEQ2SEQ (SS2S) \citep{pourjafari2022survival} addresses this issue and takes advantage of RNN cells in their framework decoder to generate smoother PDFs. However, its approach to handling censoring is overly simplistic, focusing solely on the premise that events should not occur before the censoring time, without addressing any potential implications after it. Another model that shares a similar concern in handling censoring is the Transformer-based Deep Survival Model (TDSM) \citep{hu2021transformer}. This deficiency is evident in the designs of their loss functions. The previous transformer architectures in survival analysis include TDSM and SurvTRACE \citep{wang2022survtrace}, but none of them have been extended to handle dynamic analysis.

In summary, while most existing methods could perform static analysis, only three of them could handle longitudinal data. Despite all the advancements in these recent works, there is a lack of a universal model which could jointly integrate handling miss data method, process various input formats and produce organized PDFs.

\section{Method}\label{sec3}

In this section, we introduce our formal framework \textbf{UniSurv}, which is a adaptive Transformer-based architecture for survival analysis. We assume that the available survival dataset is subject to right censoring.

\subsection{Survival Notation}

We denote time-invariant and time-varying covariates by $\pmb{x}_n$ and $\pmb{x}_v$, probability by $p$, time by $T$, $t$ or $\tau$, PDF by $p(t)$ and survival function by $S(t)$. Let's represent the survival dataset as ${(\pmb{x}_{n}^{i}, \pmb{x}_{v}^{i}, T^{i}, \delta^{i})}_{i=1}^{N}$, where for individual $i$, $\delta^{i}$ is the event indicator typically taken from the set $\{0,1\}$ without competing risks, and $T^{i}$ represents the event or censoring time depending on $\delta^{i}$. We omit the explicit dependence on $i$ throughout this and the next subsections for simplifying notation.

We assume that time $t \in \{T_{0}, T_{1}, ..., T_{max}\}$ to fit a discrete survival model, where $t$ is a discrete random variable, and $T_j$ is each time step with equal interval. The cumulative distribution function (CDF) of $t$ can be easily calculated by its PDF as
\begin{equation}
CDF(T_{j} \mid (\pmb{x}_n,\pmb{x}_v)) = \sum_{t=T_0}^{T_j}p_{t} 
\end{equation}

Having defined the probability that an event has occurred by duration $T_{j}$, the survival function can then be estimated as the probability that the survival time $t$ is at least $T_{j}$. It can also be represented as the complement of CDF as 
\begin{equation}
S(T_{j} \mid (\pmb{x}_n,\pmb{x}_v)) = 1 - CDF(T_{j} \mid (\pmb{x}_n,\pmb{x}_v)) = \sum_{t=T_j}^{T_{max}}p_{t} \label{eq_S}
\end{equation}
\subsection{Model Description}

\begin{figure}[]
\centering
\subfigure[UniSurv framework]{
\adjustbox{valign=c}{\includegraphics[width=.705\textwidth]{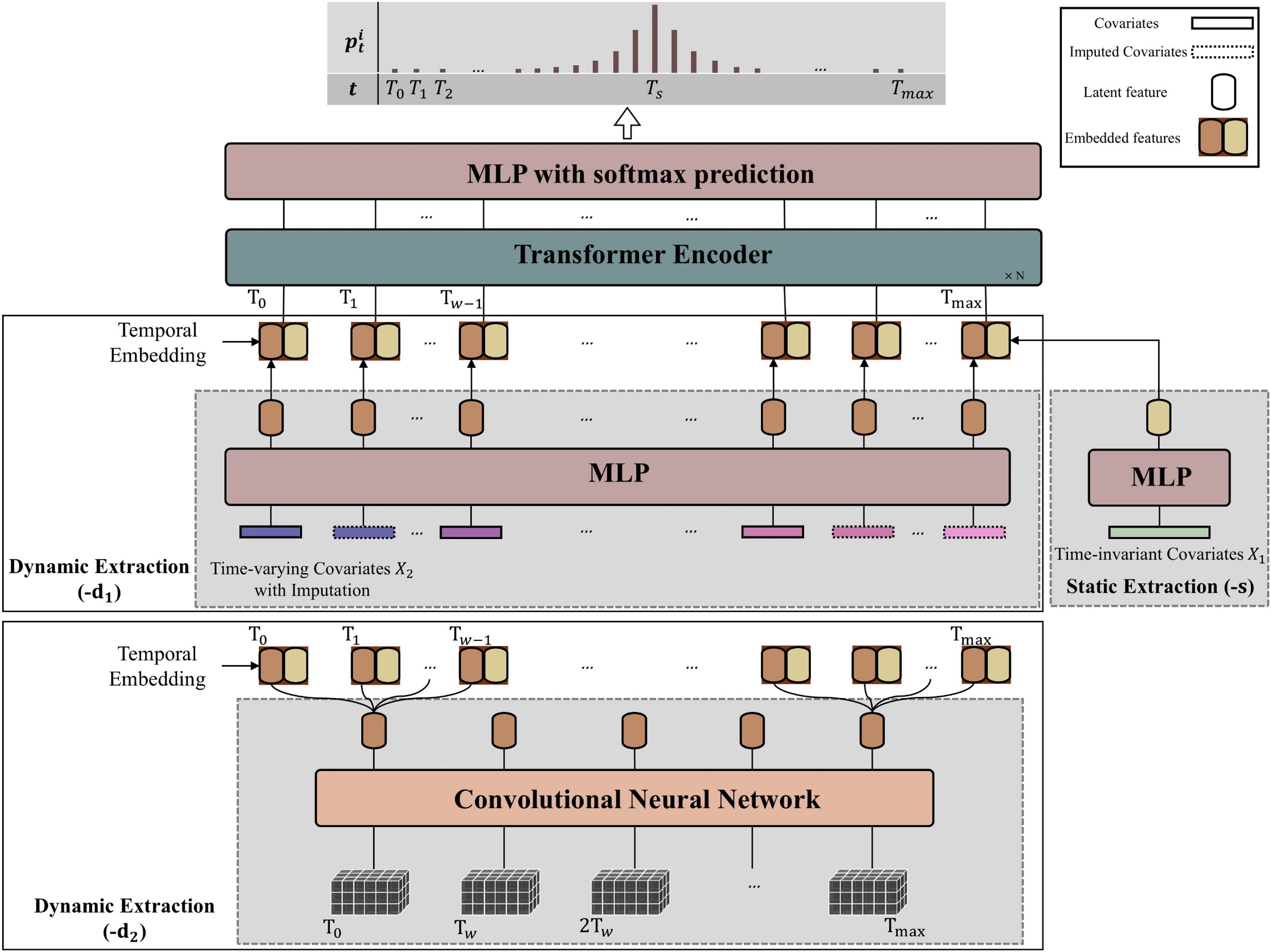}}
\label{model}
}
\subfigure[Overall schemata]{
\label{stage}
\adjustbox{valign=c}{\includegraphics[width=.25\textwidth]{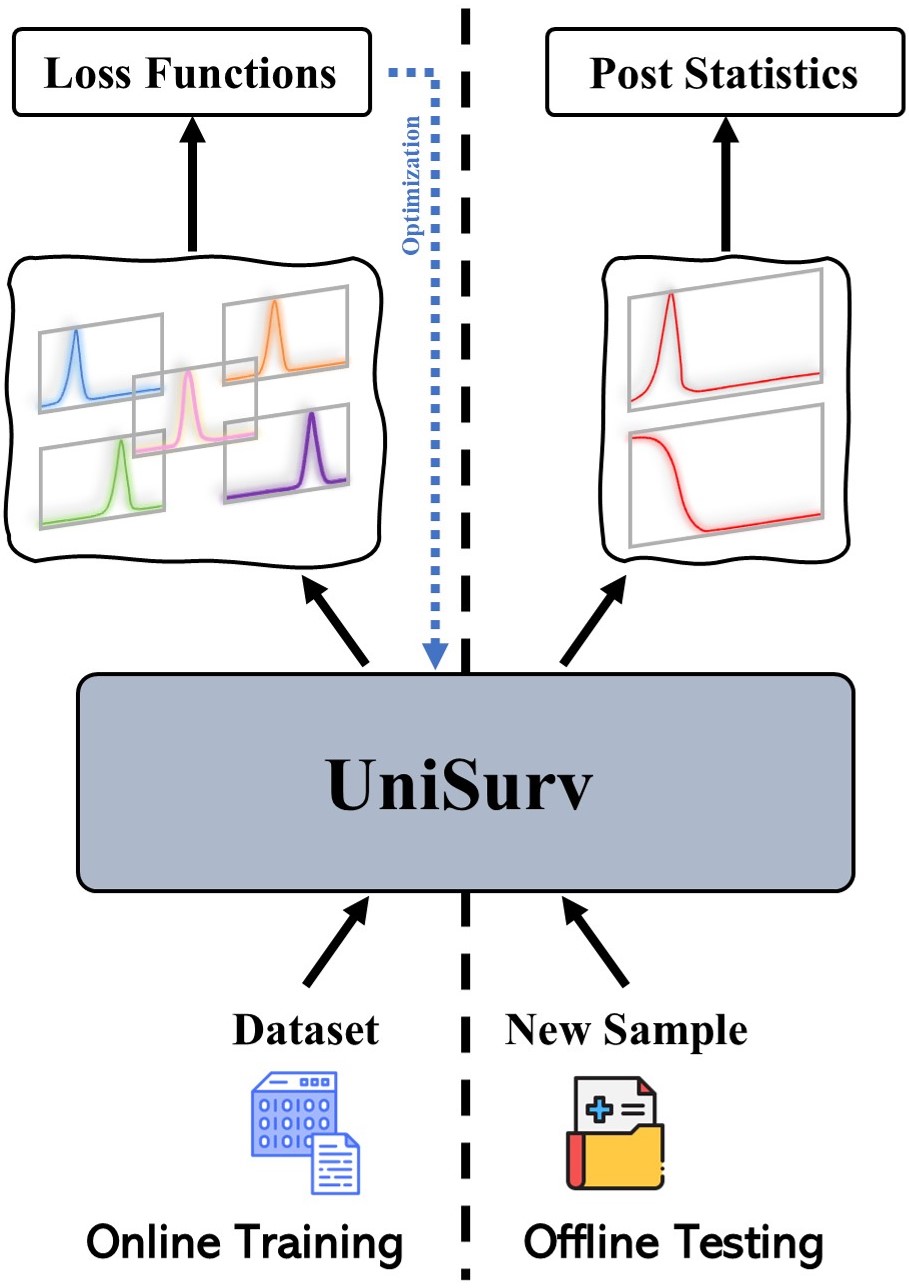}}
}
\caption{The illustration of (a) the architecture of UniSurv model and (b) a schematic representation of the UniSurv during training and testing stages}
\end{figure}

Fig. \ref{model} presents a comprehensive illustration of the UniSurv model. It integrates a novel survival loss design to enable a seamless end-to-end learning procedure. It encompasses dynamic and static extraction components, coupled with a Transformer encoder module culminating in a softmax layer at the output terminal. Besides, Fig. \ref{stage} depicts the conceptual process of UniSurv during both training and testing stages.

\subsubsection{Static and Dynamic Extractions} We integrate the variation of \textit{last-observation-carried-forward} (LOCF) method to handle missing data. It duplicates the value of the last observation to replace the following missing values until $\Delta_{\tau}=3$, and the ones that are still missing after LOCF are imputed by mean/mode of all previous points for continuous/binary covariates until $T_{max}$, making sure no missing data for all time points $T_j$ \citep{lee2019dynamic}.
Next, we extract latent representations of time-invariant features $\pmb{x}_n$ and time-varying features $\pmb{x}_v$ by static (-s) and tabular-data-based dynamic (-d{\tiny 1}) extraction modules separately. These modules are constructed using MLPs to deal with numerical tabular data. The representation of $\pmb{x}_n$ is replicated $T_{max}+1$ times by encompassing $T_0$ as well. For static modelling, these are subsequently transmitted to encoder. For dynamic modelling, these representations are concatenated with the representation of $\pmb{x}_v$ at their respective time points before the transmission. Meanwhile, a convolutional neural network (CNN) variation (-d{\tiny 2}) of the image-based dynamic extraction module can be used to address the image-like input formats. As shown in Fig. \ref{model}, the extracted feature can be shared across a predefined time window $T_w$ in light of the data sparsity.

\subsubsection{Transformer Encoder}
The core of the encoder module is a Transformer, which treats each patient as a `sentence' and the embedded features as `words' of the sentence. For an input sample, the number of words correspond to the duration $t \in \{0,1,...,T_{max}\}$, where we predefined $T_0=0$ and $T_{max}$ is a hyper-parameter selected based on the longest temporal data of a dataset. The previous concatenated representations pass through a MLP followed by layer normalization to get the embedded features. Following the conventional approach of a Transformer, we utilize the sine-cosine positional embedding \citep{vaswani2017attention} as temporal embedding in this work, and add it onto the set of embedded features, whose length is set as the embedding dimension $d_{m}$ of the Transformer. The Transformer encoder then processes embedded features and produces $T_{max}+1$ outputs, each with shape $1 \times d_{m}$.
It is worth noting that the self-attention layers in the encoder is modified to prevent positions from attending to subsequent positions. Specifically, it prohibits each position from attending to subsequent positions, and the attention scores for all illegal connections are masked out by assigning them with $-\infty$ \citep{vaswani2017attention}.
Next, all-time point outputs are fed into an exclusive 2-layer MLP. The first layer is followed by rectified linear unit (ReLU) and layer normalization,  with shape of $d_{m} \times \frac{d_{m}}{2}$. The second layer, with shape of $d_{m} \times 1$, is followed by a softmax layer to produce the individual estimated PDF. Further, the estimated survival function $\hat{S}(t \mid (\pmb{x}_n,\pmb{x}_v))$ can be calculated based on Eq. \ref{eq_S}, which ensures its monotonicity is preserved. Moreover, in discrete survival analysis, the mean lifetime $\mu$ can be approximated by the sum of the survival probabilities up to $T_{max}$. We could get the estimated mean lifetime by further involving Eq. \ref{eq_S} as
\begin{equation}
\hat{\mu} = \sum_{t=T_0}^{T_{max}}\hat{S}(t)=\sum_{t=T_0}^{T_{max}}\sum_{\tau=t}^{T_{max}} \hat{p}_{\tau}
\approx  \sum_{t=T_0}^{T_{max}} t \cdot \hat{p}_{t}
\end{equation}
where the employment of the approximately equal symbol in the equation is attributed to the presence of time point $T_0=0$. The variance of distribution is computed as
\begin{equation}
v = \sum_{t=T_0}^{T_{max}} \hat{p}_{t}\cdot (t-\hat{\mu})^{2}
\label{variance}
\end{equation}

\subsection{Loss Function}
To robustly estimate the uncensored survival time via distribution learning and generating smooth PDF, we adopt a variation of Mean-Variance loss \citep{pan2018mean} in UniSurv, which requiring each training sample has a corresponding event time label. However, censoring existing in survival dataset does not have event but censoring time. Using censoring time as the label will be misleading for the model, resulting in prediction bias. To overcome this, we employ the ``margin time'' concept \citep{haider2020effective}, to assign a ``best guess'' value to each censored subject based on the non-parametric population Kaplan-Meier (KM) \citep{kaplan1958nonparametric} estimator. Given a subject $i$ censored at time $T^{i}$, its margin event time is calculated by
\begin{equation}
e_{m}^{i} = T^{i} + \frac{\int_{T^{i}}^{T_{max}}S_{km(D_{tr})}(t)dt}{S_{km(D_{tr})}(T^{i})}
\end{equation}
where $S_{km(D_{tr})}$ is the KM estimation derived from the training dataset. It is worth to mention that during the integration process, we compute it up to $T_{max}$ in this work, ensuring that $e_{m}$ remains within the bounds of $T_{max}$, which stands in contrast to the approach proposed by \cite{haider2020effective}. They extend the KM curve infinitely through risky extrapolation beyond observed values.

We denote $T^{i}$ as the corresponding ground-truth event/censoring time for individual $i$. With the estimated mean lifetime $\hat{\mu}^{i}$, the Margin-Mean loss can be computed as
\begin{equation}
\mathcal{L}_{mm} = \frac{1}{2} \sum_{i=1}^{N}\Big{(}\delta^i \cdot  (\hat{\mu}^i - T^i)^2 + (1-\delta^i) \cdot \omega^i \cdot (\hat{\mu}^i - e_{m}^i)^2\Big{)}
\label{lm}
\end{equation}
where $\omega^i = 1 - S_{km(D_{tr})}(T^i)$ can give high confidence weight with late censor time but low with early censor time. Margin-Mean loss can penalize dissimilarity between estimated mean lifetime and actual event/margin-event time. Besides, the Variance loss is calculated as
\begin{equation}
\mathcal{L}_{v} = \sum_{i=1}^{N}v^{i}
\end{equation}
which is implemented to regulate the spread of the estimated survival distribution, limiting it to a narrow range within the mean. Considering with Eq. \ref{variance}, $\mathcal{L}_{v}$ can cause the probabilities at time points farther from $\hat{\mu}^{i}$ to approach $0$. The softmax loss, as known as cross-entropy loss, can be computed as
\begin{equation}
\mathcal{L}_{s} = \sum_{i=1}^{N}-logp_{i, T^i}
\end{equation}
which is further utilized to aid in early training convergence, as Margin-Mean-Variance loss alone may experience substantial fluctuations \citep{pan2018mean}.

Finally, tailored to address uncensoring, we utilize discordant loss by
\begin{equation}
\mathcal{L}_{d} = \sum_{i=1}^{N}\delta^i \cdot max\{0, (T^{k}-T^{i})-(\hat{\mu}^k-\hat{\mu}^i)\} \label{eq_Ld}
\end{equation}
which can penalize the randomized discordant pairs for improving model's pairwise ranking ability. The process is similar to randomized algorithm: random sampling with replacement individual $k$ for each confirmed individual $i$, making sure that $T^{k}>T^{i}$ and the difference between estimated times should not be smaller than the difference between ground truths. $\mathcal{L}_{d}$ can further penalize the discordant pairs because when $T^k$ and $T^i$ are close, the Margin-Mean-Variance loss cannot effectively discriminate discordant pairs and may fall into a local optimum.

The total loss to train UniSurv is the combination of the above four losses as
\begin{equation}
\mathcal{L}_{total} = \mathcal{L}_{s} + \lambda_{m}\mathcal{L}_{mm} + \lambda_{v}\mathcal{L}_{v} + \lambda_{d}\mathcal{L}_{d}
\end{equation}
where $\lambda_{m}$, $\lambda_{v}$, $\lambda_{d}$ are weights for the corresponding loss functions.

\section{Experiments}\label{sec4}

In this section, we demonstrate the effectiveness of UniSurv by comparing it with other benchmarks on real and synthetic datasets from static and dynamic settings.

\subsection{Datasets}
To highlight the right-skewed characteristic of survival data, we utilized three real-world datasets and two long-tailed synthetic datasets.
\subsubsection{Static Datasets}
The Study to Understand Prognoses Preferences Outcomes and Risks of Treatment (\textbf{SUPPORT}) \citep{knaus1995support} is a large static survival dataset of seriously ill hospitalized adults.
The Molecular Taxonomy of Breast Cancer International Consortium (\textbf{METABRIC}) \citep{curtis2012genomic} is a static breast cancer dataset aiming to distinguish its subtypes based on the molecular characteristics.
Their pre-processing strategies follow DeepSurv \citep{katzman2018deepsurv}.

We also generate a static synthetic dataset (\textbf{SYNTH-s}) of the style of that in \cite{lee2018deephit} but without competing risks. The dataset contains $N=15,100$ examples drawn from the stochastic process
\begin{equation}
\begin{split}
&\pmb{x}_{n}^{i} \sim \mathcal{N} (0, \mathbf{I})\\
&T^{i} \sim {\rm exp}(\pmb{\gamma}_{n}^{T}\pmb{x}_{n}^{i})
\end{split}
\label{synths_process}
\end{equation}
where $\pmb{x}_{n}^{i}$ is a vector of 4-dimensional variables and $\pmb{\gamma}_{n}=\pmb{10}$. We randomly select $50\%$ patients to be right-censored with random censoring time uniformly drawn from $[0, T^{i}]$. More details are listed in Tab. \ref{data_info}.

\begin{table*}[]
\centering
\caption{Descriptive statistics of three real world medical datasets and two synthetic datasets}
\resizebox{1\textwidth}{!}{%
\begin{tabular}{c|c|cc|cc|ccc|ccc}
\hline
\multirow{2}{*}{\textbf{Dataset}} & \multirow{2}{*}{\textbf{Longitudinal}} & \multirow{2}{*}{\textbf{Uncensored}} & \multirow{2}{*}{\textbf{Censored}} & \multicolumn{2}{c|}{\textbf{Features}} & \multicolumn{3}{c|}{\textbf{Event Time}} & \multicolumn{3}{c}{\textbf{Censoring Time}} \\
 &  &  &  & \textbf{static} & \textbf{dynamic} & \textbf{min} & \textbf{max} & \textbf{mean} & \textbf{min} & \textbf{max} & \textbf{mean} \\ \hline
\textbf{SUPPORT} & No & 6036 (68\%) & 2837 (32\%) & 14 & - & 0 & 65 & 6.85 & 12 & 68 & 35.33 \\ \hline
\textbf{METABRIC} & No & 1103 (58\%) & 801 (42\%) & 9 & - & 0 & 355 & 99.95 & 0 & 337 & 159.55 \\ \hline
\textbf{SYNTH-s} & No & 7600 (50\%) & 7500 (50\%) & 4 & - & 0 & 192 & 22.45 & 0 & 165 & 10.80 \\ \hline \hline
\textbf{MSReactor} & Yes & 148 (20\%) & 598 (80\%) & 8 & 90 & 1 & 68 & 17.54 & 0 & 80 & 45.43 \\ \hline
\textbf{SYNTH-d} & Yes & 7462 (49\%) & 7638 (51\%) & 4 & 20 & 0 & 199 & 57.96 & 0 & 195 & 28.59 \\ \hline
\end{tabular}%
}
\label{data_info}
\end{table*}

\subsubsection{Dynamic Datasets}
On the basis of SYNTH-s, we further generate dynamic synthetic dataset (\textbf{SYNTH-d}) by adding additional dynamic variables following Weibull distribution, and introducing temporal noise disturbances to make them variable over time as
\begin{equation}
\begin{split}
&\pmb{x}_{v}^{i}(t) \sim \frac{a}{\beta}\Big(\frac{\pmb{x}}{\beta }\Big)^{(a-1)}exp\Big(-(\frac{\pmb{x}}{\beta})^a\Big) + \mathcal{N} (0, \mathbf{I})\\
&T^{i} \sim {\rm exp}\Big(\pmb{\gamma}_{v_{1}}^{T} \cdot max\big(\pmb{x}_{v_{1}}^{i}(t)\big)+\pmb{\gamma}_{v_{2}}^{T} \cdot min\big(\pmb{x}_{v_{2}}^{i}(t)\big)+\pmb{\gamma}_{n}^{T}\pmb{x}_{n}^{i}\Big)
\end{split}
\label{synthd_process}
\end{equation}
where $\pmb{x}_{v}^{i}$ is a $20 \times T_{max}$ dynamic variable matrix for all time points, $a$ is the shape parameter, $\beta$ is the scale parameter, $\pmb{\gamma}_{v_{1}}=\pmb{\gamma}_{v_{2}}=\pmb{5}$, and $max(\cdot)$ and $min(\cdot)$ are operations on the temporal dimension. Besides, $v_{1}$ and $v_{2}$ are two randomly selected subsets that satisfy $v_{1} \cap v_{2} = \varnothing$, $v_{1} \cup v_{2} = v$. $T^{i}$ is then resampled, and the method  of introducing censoring cases remains the same as before.

Moreover, \textbf{MSReactor} \citep{merlo2021association} dataset is a quantifiable, objective collection on cognition via longitudinal computerized test for Multiple Sclerosis (MS), integrating with other 8 static covariates.
In each test, patients are instructed to respond as quickly as possible to onscreen stimuli, and their reaction time is recorded in millisecond (ms). The test includes 3 different tasks for testing their psychomotor function, attention and working memory. Each patient undergoes the test a number of times after the diagnosis and prior to the occurrence of the event/censoring (with at least one-month interval between every two adjacent tests). The survival event is characterized by EDSS progression through the \texttt{six-month disability worsening confirmation rule} \citep{hunter2021confirmed}. Numerous research investigations have indicated that utilizing reaction data could potentially offer a more responsive approach for detecting subclinical cognitive impairment in comparison to current cognitive assessment methods \citep{foong2023smartphone,pham2021smartphone,whitehouse2019comorbid}.

\begin{figure}[]
\centering
\includegraphics[width=1\columnwidth]{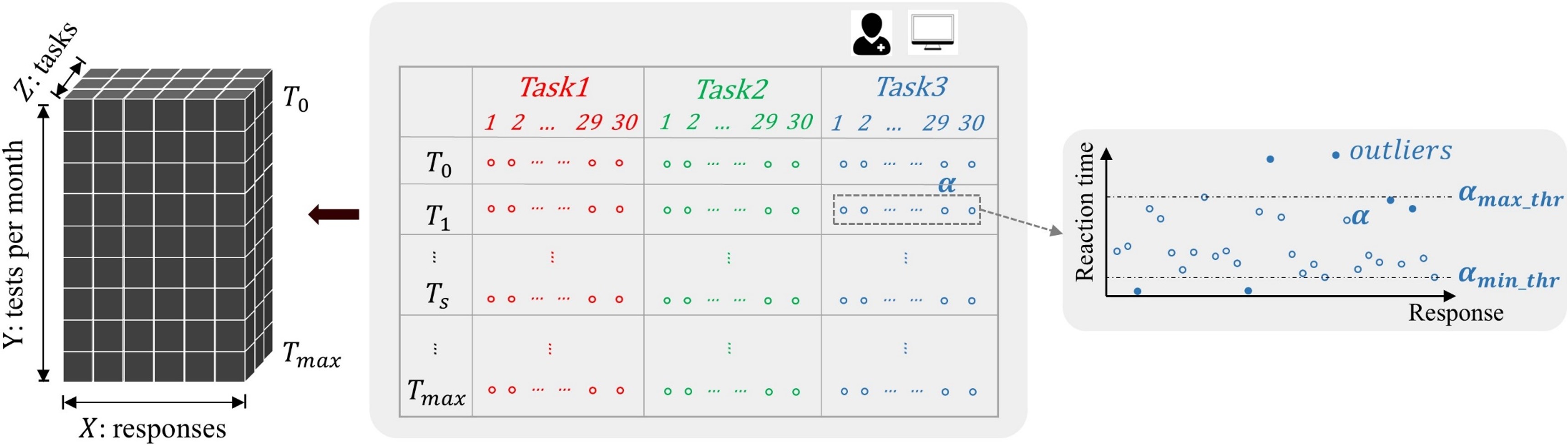}
\caption{The illustration of reaction tensor representation of a single individual in MSReactor}
\label{data_rep}
\end{figure}

The longitudinal reaction test will be considered as time-varying covariates. However, due to certain redundancies present in MSReactor, evident through pronounced inter-column associations, the characteristics of adjacent columns exhibit robust correlations, deviating from the conventional tabular data extraction where each column represents highly streamlined information. Without the application of specialized data preprocessing techniques and innovative model architectures, the existing survival models may encounter difficulties in extracting meaningful latent patterns from this data. Therefore, we transform the longitudinal tabular data part into a composite ``\textit{reaction tensor}" after monthly imputation, and utilize the certain module to deal with it. Specifically, each patient has a unique 3-dimensional reaction tensor\footnote{More details are illustrated in Appendix \ref{secA2}.}, as shown in Fig. \ref{data_rep}. Its Z-axis is corresponding to the 3 different tasks. X-axis is response dimension with fixed length of 30, corresponding to the 30 times the patient needs to finish three tasks separately in each test. Y-axis corresponds to the times patient has undergone tests per month with fixed length from the start time $T_0$ to the end time $T_{max}$. The reaction tensor is divided into several smaller tensors along Y-axis by $T_w$ as in Fig. \ref{model}.

\subsection{Evaluation Metrics}
We utilize ranking measures such as concordance index (\textbf{C-index}) from \texttt{lifelines} \citep{Davidson-Pilon2019} library and mean cumulative area under ROC curve (\textbf{mAUC}) from \texttt{scikit-survival} \citep{sksurv} library, and accuracy measures such as mean absolute error (\textbf{MAE}) as the evaluation metrics for all experiments.

\subsubsection{Concordance}
C-index \citep{uno2011c} is able to estimate ranking ability by comparing relative risks across all pairs in the test set as
\begin{equation}
\text{C-index} = \frac{ {\textstyle \sum_{i, k}} \delta^i \cdot \mathbb{I}(T^{i}<T^{k})\cdot \mathbb{I}(\hat{\mu}^{i}<\hat{\mu}^{k})}
{{\textstyle \sum_{i, k}} \delta^i \cdot \mathbb{I}(T^{i}<T^{k})} 
\end{equation}
where $\mathbb{I}(\star)$ is an indicator function, and $\delta_i=0$ if $T^i$ is uncensored and 1 otherwise.

\subsubsection{MAE-Uncensored}
MAE-Uncensored (MAE-U) can compensate for the inability of C-index to measure the mean absolute value of the estimated risk score. It is computed as 
\begin{equation}
\text{MAE-U} = \frac{{\textstyle \sum_{i}} \delta^i \cdot \left | T^i - \hat{\mu}^i \right |}{{\textstyle \sum_{i}} \delta^i}
\end{equation}

\subsubsection{MAE-Hinge}
MAE-Hinge (MAE-H) is a one-sided MAE for only censoring cases, opposite with MAE-U for uncensoring only. It considers only if the predicted time $\hat{\mu}$ is earlier than the censored time $T$ as follow
\begin{equation}
\text{MAE-H} = \frac{{\textstyle \sum_{i}} (1-\delta^i) \cdot max\{ T^i - \hat{\mu}^i , 0\}}{{\textstyle \sum_{i}} (1-\delta^i)}
\end{equation}

\subsubsection{Mean Cumulative Area Under ROC Curve}
The area under ROC curve for survival analysis involves treating survival issue as binary classification across various quantiles of event times and defining the sensitivity and specificity as time-dependent measures \citep{lambert2016summary}. The cumulative AUC measures model's capability of discriminating individuals who fail by a specified $t$ ($T_j\leq t$) from subjects who fail after this time ($T_j > t$). We compute the mAUC by integrating the cumulative AUC over all time range $(T_j, T_j+1)$.

\subsection{Experimental Setting}
We compare with five static benchmarks, including \textbf{CPH}, \textbf{DeepSurv}, \textbf{DeepHit}, \textbf{DSM} and \textbf{TDSM}, and two dynamic benchmarks\footnote{We have not compared with SS2S in this study as its code has not been made publicly at the moment.}, including \textbf{DDH} and \textbf{RDSM}. As static dataset does not have longitudinal covariates, our dynamic extraction module in UniSurv is in non-activation mode named UniSurv-s.
For MSReactor, the dynamic extraction module has two variants based on different data representations, tabular data representation named UniSurv-d{\tiny 1} and image-like representation named UniSurv-d{\tiny 2}. More implementation and hyperparameter details are in the Appendix \ref{secA4}\footnote{Code availability: \href{https://github.com/XinZ0419/UniSurv}{https://github.com/XinZ0419/UniSurv}}.

For a fair comparison, we use C-index as early stopping criterion for all approaches as it can cover more subjects than MAE. We report the results by using cross-validation, randomly splitting datasets 5 times into training, validation and test sets with ratio 7:1:2. All experiments are implemented in PyTorch 2.0.1 on the same environments with a fixed random seed.

\subsection{Benchmarking Results}
\begin{table*}[htbp]
\centering
\caption{Benchmarking on three static datasets. ``\dag'' denotes P-Value $<0.05$, where “w/o mask” means "without masking" and is not in comparison. Higher ($\uparrow$) values of C-index and mAUC; and lower ($\downarrow$) values of MAE-U and MAE-H are better}
\resizebox{\textwidth}{!}{%
\begin{tabular}{ccccccccccccc}
\hline
\multicolumn{1}{c|}{\multirow{2}{*}{\textbf{Model}}} & \multicolumn{4}{c|}{\textbf{SUPPORT}} & \multicolumn{4}{c|}{\textbf{METABRIC}} & \multicolumn{4}{c}{\textbf{SYNTH-s}} \\ \cline{2-13} 
\multicolumn{1}{c|}{} & \textbf{C-index} $\uparrow$ & \textbf{MAE-U} $\downarrow$ & \textbf{MAE-H} $\downarrow$ & \multicolumn{1}{c|}{\textbf{mAUC} $\uparrow$} & \textbf{C-index} $\uparrow$ & \textbf{MAE-U} $\downarrow$ & \textbf{MAE-H} $\downarrow$ & \multicolumn{1}{c|}{\textbf{mAUC} $\uparrow$} & \textbf{C-index} $\uparrow$ & \textbf{MAE-U} $\downarrow$ & \textbf{MAE-H} $\downarrow$ & \textbf{mAUC} $\uparrow$ \\ \hline
\multicolumn{1}{c|}{\textbf{CPH}} & $0.585_{.012}$ & $19.25_{1.23}$ & $17.85_{3.12}$ & \multicolumn{1}{c|}{$0.715_{.013}$} & $0.633_{.023}$ & $81.17_{4.13}$  & $33.27_{2.10}$  &  \multicolumn{1}{c|}{$0.808_{.021}$} & $0.702_{.007}$ & $22.85_{3.36}$  & $5.48_{0.25}$ & $0.838_{.007}$ \\ \hline
\multicolumn{1}{c|}{\textbf{DeepSurv}} & $\pmb{0.610}_{.017}$ & $18.76_{1.12}$ & $\underline{14.93}_{2.58}$ &  \multicolumn{1}{c|}{$\pmb{0.789}_{.016}$} & $\pmb{0.642}_{.020}$ & $77.68_{5.12}$ & $34.81_{1.85}$ &  \multicolumn{1}{c|}{$0.822_{.016}$} & $0.701_{.009}$ & $21.74_{3.06}$ & $5.85_{0.46}$ & $0.835_{.008}$ \\ \hline
\multicolumn{1}{c|}{\textbf{DeepHit}} & $0.601_{.014}$ & $17.45_{2.57}$ & $21.25_{3.43}$ &  \multicolumn{1}{c|}{$0.738_{.018}$} & $0.636_{.021}$ & $78.25_{5.14}$ & $\underline{31.65}_{1.75}$ &  \multicolumn{1}{c|}{$0.811_{.018}$} & $\underline{0.723}_{.006}$ & $22.37_{3.19}$ & $5.45_{0.14}$ & $\underline{0.859}_{.005}$ \\ \hline
\multicolumn{1}{c|}{\textbf{DSM}} & $0.602_{.005}$ & $17.58_{2.19}$ & $19.69_{3.74}$ &  \multicolumn{1}{c|}{$0.742_{.007}$} & $0.633_{.028}$ & $75.18_{4.31}$ & $32.21_{2.13}$ &  \multicolumn{1}{c|}{$0.805_{.011}$} & $0.685_{.010}$ & $24.17_{3.74}$ & $\underline{5.36}_{0.26}$ & $0.823_{.010}$ \\ \hline
\multicolumn{1}{c|}{\textbf{TDSM}} & $0.603_{.007}$ & $\pmb{8.67}^{\dag}_{0.73}$ & $29.32_{3.41}$ &  \multicolumn{1}{c|}{$0.762_{.011}$} & $0.637_{.018}$ & $\pmb{55.97}^{\dag}_{5.83}$ & $62.33_{2.91}$ &  \multicolumn{1}{c|}{$\underline{0.824}_{.017}$} & $0.718_{.007}$ & $\pmb{11.25}_{3.57}$ & $6.47_{0.34}$ & $0.850_{.006}$ \\ \hline \hline
\multicolumn{1}{c|}{\textbf{UniSurv-s}} & $\underline{0.604}_{.007}$ & $\underline{17.35}_{1.55}$ & $\pmb{12.92}^{\dag}_{2.61}$  &  \multicolumn{1}{c|}{$\underline{0.767}_{.012}$} & $\underline{0.638}_{.021}$ & $\underline{71.30}_{6.23}$ & $\pmb{23.06}^{\dag}_{2.68}$ &  \multicolumn{1}{c|}{$\pmb{0.826}_{.014}$} & $\pmb{0.731}_{.008}$ & $\underline{19.15}_{3.23}$ & $\pmb{1.55}^{\dag}_{0.04}$  & $\pmb{0.866}_{.006}$ \\ \hline
\multicolumn{1}{c|}{\textbf{UniSurv-s} \footnotesize{w/o mask}} & $0.603_{.008}$ & $17.42_{1.56}$ & $12.80_{2.63}$ &  \multicolumn{1}{c|}{$0.769_{.010}$} & $0.638_{.020}$ & $73.22_{5.45}$ & $22.32_{2.29}$ &  \multicolumn{1}{c|}{$0.826_{.014}$} & $0.732_{.007}$ & $19.23_{3.28}$ & $1.51_{0.04}$ & $0.868_{.005}$ \\ \hline
\multicolumn{1}{l}{} & \multicolumn{1}{l}{} & \multicolumn{1}{l}{} & \multicolumn{1}{l}{} & \multicolumn{1}{l}{} & \multicolumn{1}{l}{} & \multicolumn{1}{l}{} & \multicolumn{1}{l}{} & \multicolumn{1}{l}{} & \multicolumn{1}{l}{} & \multicolumn{1}{l}{} & \multicolumn{1}{l}{} & \multicolumn{1}{l}{} \\
\multicolumn{1}{l}{} & \multicolumn{1}{l}{} & \multicolumn{1}{l}{} & \multicolumn{1}{l}{} & \multicolumn{1}{l}{} & \multicolumn{1}{l}{} & \multicolumn{1}{l}{} & \multicolumn{1}{l}{} & \multicolumn{1}{l}{} & \multicolumn{1}{l}{} & \multicolumn{1}{l}{} & \multicolumn{1}{l}{} & \multicolumn{1}{l}{}
\end{tabular}%
}
\label{result_s}
\end{table*}

Performance comparisons for all datasets are summarized in Tab. \ref{result_s} and Tab. \ref{result_d}. We bold the \textbf{best} and underline the \underline{second best}. Besides, the statistical significance is determined by paired t-test between the best results and all others individually.

\subsubsection{Static Modelling Results}
In terms of C-index, our UniSurv-s secures the first position on SYNTH-s and the second position on both SUPPORT and METABRIC. It also reaches the best mAUC on METABRIC and SYNTH-s and the second best on SUPPORT. DeepSurv shows comparable ranking performance on two real-world datasets. This illustrates that parametric model still hold a slight advantage over non-parametric model, rely on its robust probability distribution assumptions. Meanwhile, the performances of the other four models vary, creating a competitive landscape. This makes it difficult to definitively judge their performance under single ranking metrics.

\begin{figure}[]
\centering
\includegraphics[width=.7\columnwidth]{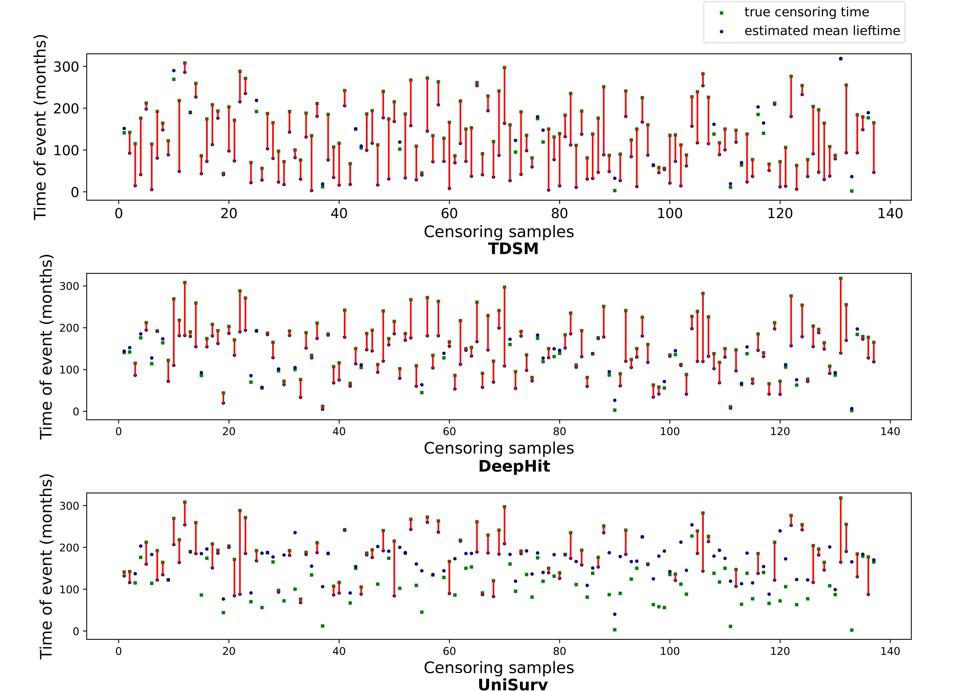}
\caption{The difference between the estimated lifetime $\hat{\mu}^i$ (blue dot) and the true censoring time $T^i$ (green square) of TDSM, DeepHit and UniSurv-s in METABRIC. Each red line indicates the difference if $\hat{\mu}^i<T^i$ for individual $i$, which is conversely not displayed in the opposite scenario}
\label{censoring_case}
\end{figure}

Meanwhile, our UniSurv performs well in MAE-U and exhibits notably superior performance in the realm of MAE-H, with statistical significance compared to other models. Only DeepSurv in SUPPORT is comparable with ours in both two MAEs. Conversely, the performance of TDSM, while excelling in MAE-U, lags notably behind in MAE-H. This is because the loss design of TDSM leads to overfitting on uncensored data throughout the learning process, failing to capture the fact that most censored samples have longer survival times. Further, the inadequacy of TDSM's predictions for censoring is also evident by Fig. \ref{censoring_case}, in which we represent the difference between true censoring time and estimated mean lifetime with red lines for some censoring cases. We show the METABRIC results from TDSM, UniSurv-s and the second-best MAE-U model DeepHit here. The more and longer red lines, the model have less sensitivity of censoring prediction. It can be observed that UniSurv has the capability to provide accurate predictions for the majority of censoring cases. This outcome can be attributed to the incorporation of the MAE-margin concept within the Margin-Mean loss $\mathcal{L}_{mm}$ in Eq. \ref{lm}, as it leverages prior knowledge from the training dataset to effectively ``enforce'' predicted survival time to exceed the censoring time. On the other hand, DeepHit exhibits significant inefficiency in forecasting longer censoring times. Similar to TDSM, this is also due to the absence of certain constraints within its loss designs beyond the censoring time, which may give rise to a systemic bias in predicting censoring cases.

\subsubsection{Dynamic Modelling Results}
As depicted in Tab. \ref{result_d}, UniSurv-d{\tiny 1} demonstrates superior performance over two other models for longitudinal datasets, as evidenced by higher values in C-index, mAUC and lower values in two MAEs. However, the performance of these three methods is generally suboptimal, as their C-index values remain below $0.6$. This occurrence likely arises from the fact that the temporal data in MSReactor diverges from conventional survival tabular data, instead representing a reaction testing approach applied to MS patients. Traditional models struggle to effectively extract meaningful insights from this intricate and redundant information. Notably, when we preprocess the computerized test data into "reaction tensor" and employ CNN to extract latent features, the performance of UniSurv-d{\tiny 2} surpasses the others with statistically significant improvements. However, this "tensor" method has not demonstrated effectiveness for SYNTH-d, primarily due to the isotropic distribution of each variable $x_{v}$ during data generation, resulting in their mutual independence and lack of correlation.

\begin{table*}[htbp]
\centering
\caption{Benchmarking on two dynamic datasets}
\resizebox{1\textwidth}{!}{%
\begin{tabular}{ccccccccc}
\hline
\multicolumn{1}{c|}{\multirow{2}{*}{\textbf{Model}}} & \multicolumn{4}{c|}{\textbf{MSReactor}} & \multicolumn{4}{c}{\textbf{SYNTH-d}} \\ \cline{2-9} 
\multicolumn{1}{c|}{} & \textbf{C-index} $\uparrow$ & \textbf{MAE-U} $\downarrow$ & \textbf{MAE-H} $\downarrow$ & \multicolumn{1}{c|}{\textbf{mAUC} $\uparrow$} & \textbf{C-index} $\uparrow$ & \textbf{MAE-U} $\downarrow$ & \textbf{MAE-H} $\downarrow$ & \multicolumn{1}{c}{\textbf{mAUC} $\uparrow$} \\ \hline
\multicolumn{1}{c|}{\textbf{DDH}} & $0.521_{.064}$ & $38.27_{10.32}$ & $16.73_{5.36}$  & \multicolumn{1}{c|}{$0.698_{.051}$} & $0.725_{.009}$ & $31.36_{4.56}$ & $5.45_{0.36}$  & $0.822_{.007}$ \\ \hline
\multicolumn{1}{c|}{\textbf{RDSM}} & $0.527_{.077}$ & $35.25_{8.12}$ & $17.92_{7.64}$ & \multicolumn{1}{c|}{$0.714_{.055}$} & $0.703_{.004}$ & $32.18_{3.67}$ & $5.32_{0.29}$ & $0.804_{.005}$ \\ \hline \hline
\multicolumn{1}{c|}{\textbf{UniSurv-d{\tiny 1}}} & $\underline{0.547}_{.032}$ & $\underline{34.12}_{10.57}$ & $\underline{10.15}_{5.12}$  & \multicolumn{1}{c|}{$\underline{0.729}_{.048}$} & $\underline{0.737}_{.008}$ & $\underline{23.74}_{5.52}$ & $\pmb{3.40}_{0.13}$ & $\underline{0.875}_{.007}$ \\ \hline
\multicolumn{1}{c|}{\textbf{UniSurv-d{\tiny 2}}} & $\pmb{0.634}^{\dag}_{.047}$ & $\pmb{29.73}_{5.29}$ & $\pmb{6.48}^{\dag}_{3.55}$  & \multicolumn{1}{c|}{$\pmb{0.793}^{\dag}_{.046}$} & $\pmb{0.739}_{.007}$ & $\pmb{23.56}_{5.73}$ & $\underline{3.52}_{0.08}$ & $\pmb{0.876}_{.008}$ \\ \hline
\multicolumn{1}{c|}{\textbf{UniSurv-d{\tiny 1}} \footnotesize{w/o mask}} & $0.562_{.030}$ & $33.53_{11.96}$ & $11.91_{8.25}$ & \multicolumn{1}{c|}{$0.732_{.058}$} & $0.739_{.007}$ & $22.47_{5.47}$ & $3.74_{0.12}$ & $0.876_{.006}$ \\ \hline
\multicolumn{1}{c|}{\textbf{UniSurv-d{\tiny 2}} \footnotesize{w/o mask}} & $0.642_{.059}$ & $28.87_{6.67}$ & $7.13_{4.47}$ & \multicolumn{1}{c|}{$0.801_{.042}$} & $0.741_{.006}$ & $23.87_{5.37}$ & $3.13_{0.10}$ & $0.878_{.007}$ \\ \hline
\multicolumn{1}{l}{} & \multicolumn{1}{l}{} & \multicolumn{1}{l}{} & \multicolumn{1}{l}{} & \multicolumn{1}{l}{} & \multicolumn{1}{l}{} & \multicolumn{1}{l}{} & \multicolumn{1}{l}{} & \multicolumn{1}{l}{} \\
\multicolumn{1}{l}{} & \multicolumn{1}{l}{} & \multicolumn{1}{l}{} & \multicolumn{1}{l}{} & \multicolumn{1}{l}{} & \multicolumn{1}{l}{} & \multicolumn{1}{l}{} & \multicolumn{1}{l}{} & \multicolumn{1}{l}{}
\end{tabular}%
}
\label{result_d}
\end{table*}

\subsubsection{The Implication of Data Distribution}
\begin{figure}[]
\centering
\includegraphics[width=.9\columnwidth]{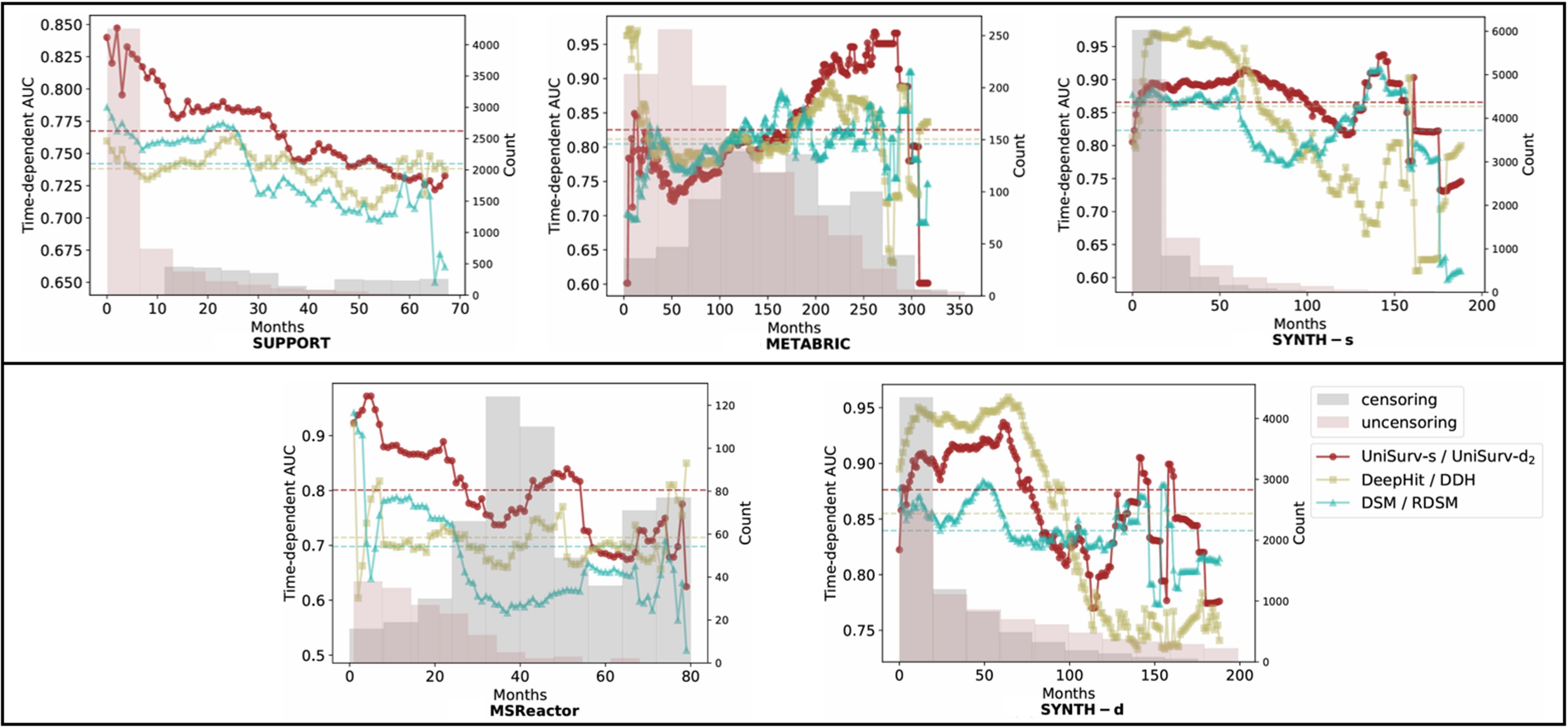}
\caption{The time-dependent AUC. The dashed line shows mAUC corresponding to each colored curve}
\label{all_auc}
\end{figure}
As shown in Fig. \ref{all_auc}, all five histograms depict the distribution of survival times skewed towards the early segment of the time horizon, while censoring times tend to cluster in the latter half, especially in SUPPORT, SYNTH-s, MSReactor and SYNTH-d. This leads to survival models facing difficulty in maintaining predictive accuracy over time, as evidenced by the time-dependent AUC (TD-AUC). For example, all the performances of UniSurv-s, DeepHit and DSM, or their dynamic variants (UniSurv-d{\tiny 2}, DDH, RDSM) exhibit a consistent decline in TD-AUC as time progresses. However, UniSurv still outperforms others, especially on two dynamic datasets. For METABRIC, due to its relatively low censoring rate and evenly distributed censoring cases, all three models maintain their TD-AUC quite well, with some even showing an upward trend, particularly UniSurv. It affirms that Transformer encoder based on Margin-Mean-Variance loss learning can effectively alleviate the challenges posed by survival datasets characterized by long-tail distributions.
\begin{figure}[]
\centering
\includegraphics[width=.7\columnwidth]{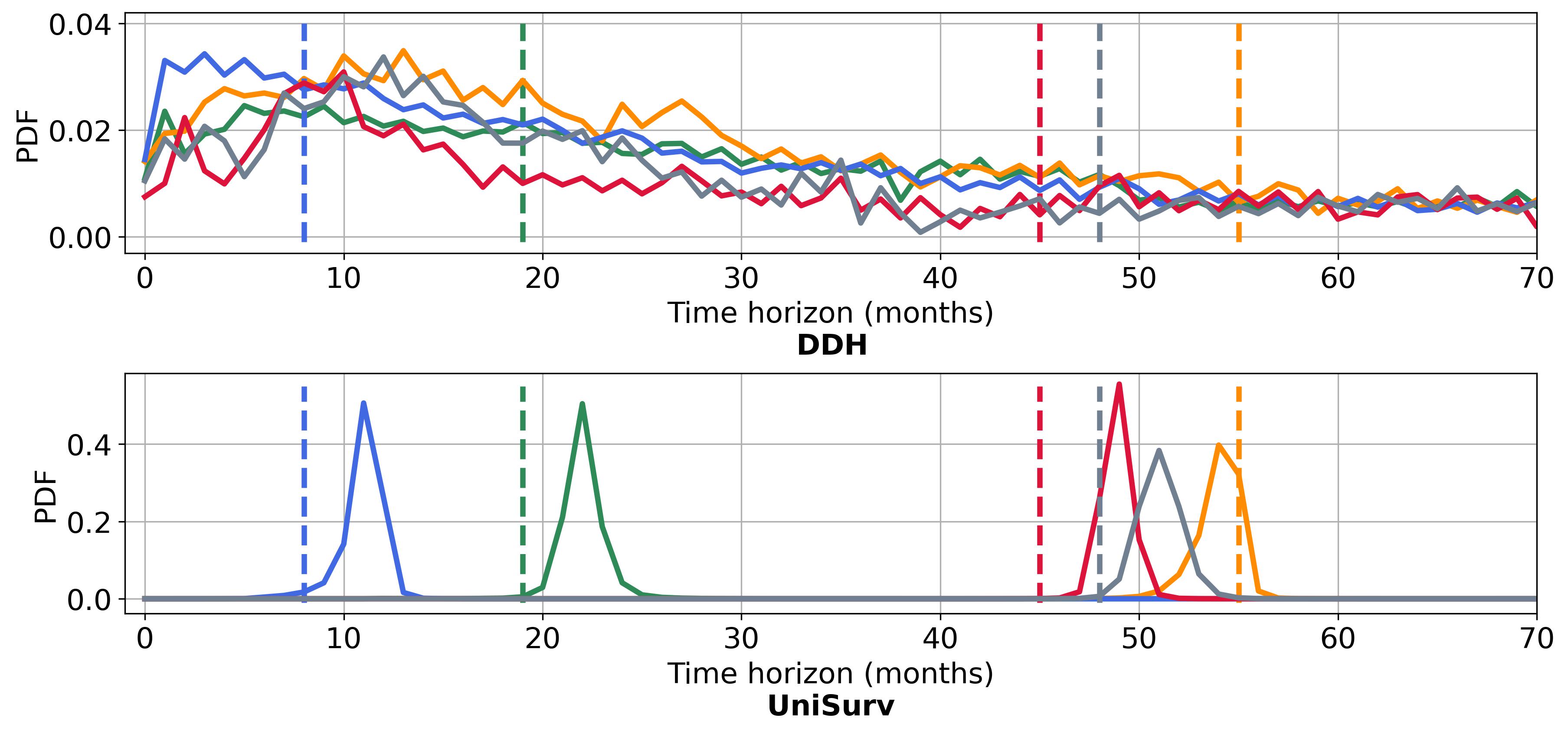}
\caption{The estimated PDFs by DDH and UniSurv for five randomly selected uncensoring cases in MSReactor. Each color represents an individual}
\label{pdf}
\end{figure}

\subsection{Importance of Masked Attention Mechanism}
In the context of leveraging Transformer for inference, the masking function within the attention mechanism is inherently discretionary, contingent upon whether each output necessitates contributions from all or specific designated inputs. For static survival data, the design of UniSurv does not entail distinctions in latent features at each time point beyond temporal embedding. Hence, there is no risk of information leakage, rendering the masking mechanism inconsequential. For instance, it is not employed in the TDSM. As demonstrated by Tab. \ref{result_s}, the overall performance of UniSurv-s has not been affected by removing masking mechanism from UniSurv-s, and the slight performance fluctuations can be negligible. However, when dealing with dynamic survival data, the missing data problem is inevitable, and imputations following event or censoring times may give rise to potential retro-active prediction concern. Therefore, the masking mechanism becomes imperative in such scenario. In Tab. \ref{result_d}, both UniSurv-d{\tiny 1} and UniSurv-d{\tiny 2} exhibited an equivalent degree of performance decline across two datasets by removing masking, which are evidenced by their ranking ability.

\subsection{Comparison of PDF Visualizations}
In addition to predictive accuracy, the quality of estimated individual PDF stands as another crucial consideration when comparing non-parametric survival models. The distribution of PDF generated by our UniSurv is specifically governed by Margin-Mean-Variance loss and remains unaffected by variations in distinct extraction modules. In Fig. \ref{pdf}, we present a comparison of the PDF outputs for 5 randomly selected uncensoring cases in MSReactor. We choose to contrast the DDH and UniSurv due to their absence of assumptions regarding the shape of the PDF, whereas RDSM relies on strong assumptions related to the Weibull and log-normal distributions. As described in above sections and shown in Fig. \ref{pdf}, our $\mathcal{L}_{mm}$ can penalize dissimilarity between the peak of PDF and the ground truth. Besides, diverging from the disordered PDFs from DDH, $\mathcal{L}_{v}$ can regulate the spread of PDF and limit it into a distinct pattern and organization. In contrast, despite using the same MLP and softmax as the output layer in DDH, the high fluctuations of PDFs can be attributed to the shortcomings in its loss function design.

The unimodal nature of survival PDF offers several advantages. For example, it can better reflect the time-to-event and naturally calibrate the median survival time corresponding to survival curve, such as \cite{rindt2022survival} employed several pre-defined unimodal distributions for survival modelling. However, UniSurv departs from this assumption, achieving the same objective through a distinctive loss design. The current over-concentrated PDF is not optimal, and appropriately adjusting $\mathcal{L}_{v}$ to relax its constraints on the shape will become necessary.

\begin{figure*}[]
\centering
\subfigure[Loss combination]{
\label{loss_ablation}
\includegraphics[width=.225\textwidth]{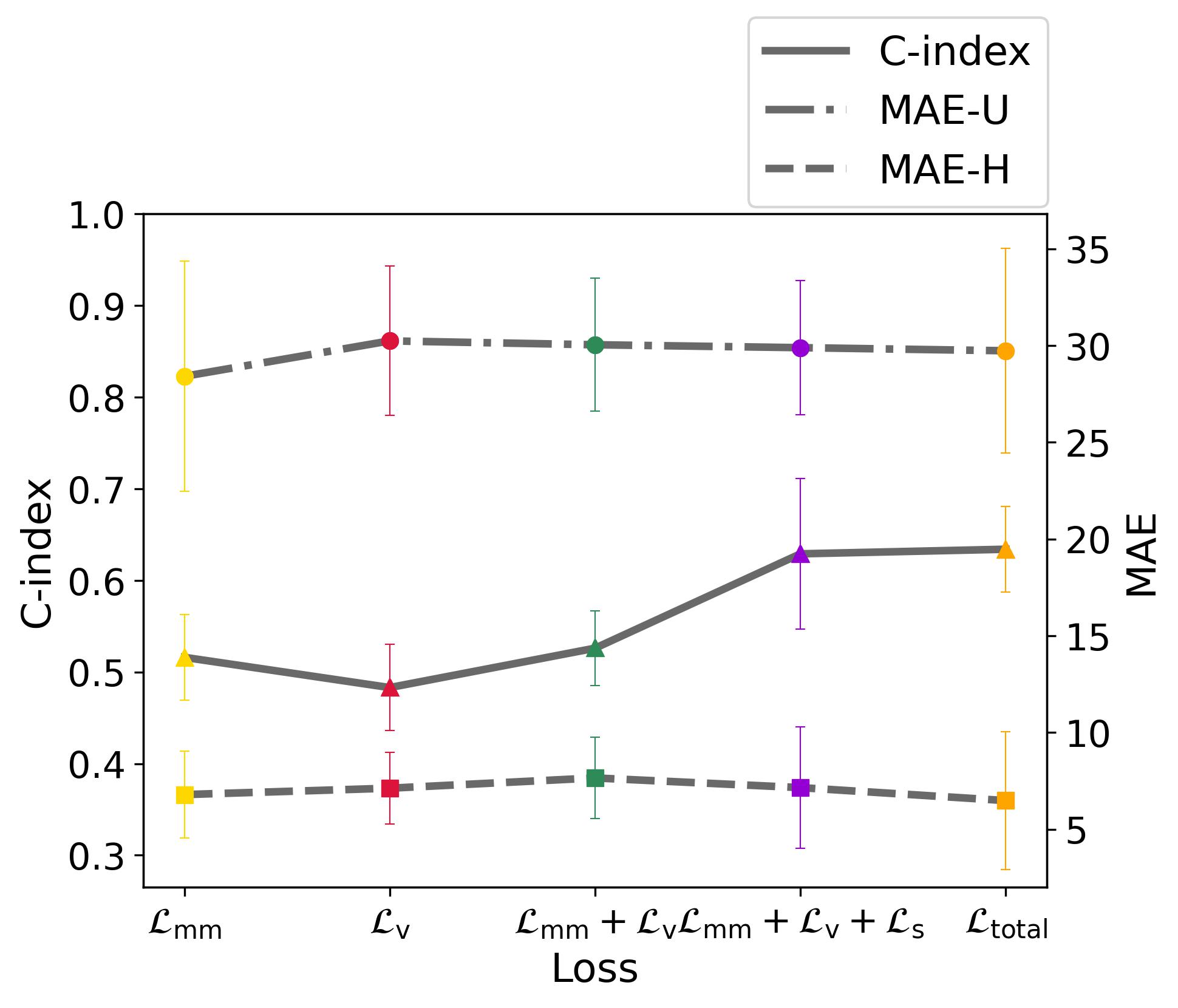}
}
\subfigure[$T_w$ selection]{
\label{Tw_ablation}
\includegraphics[width=.225\textwidth]{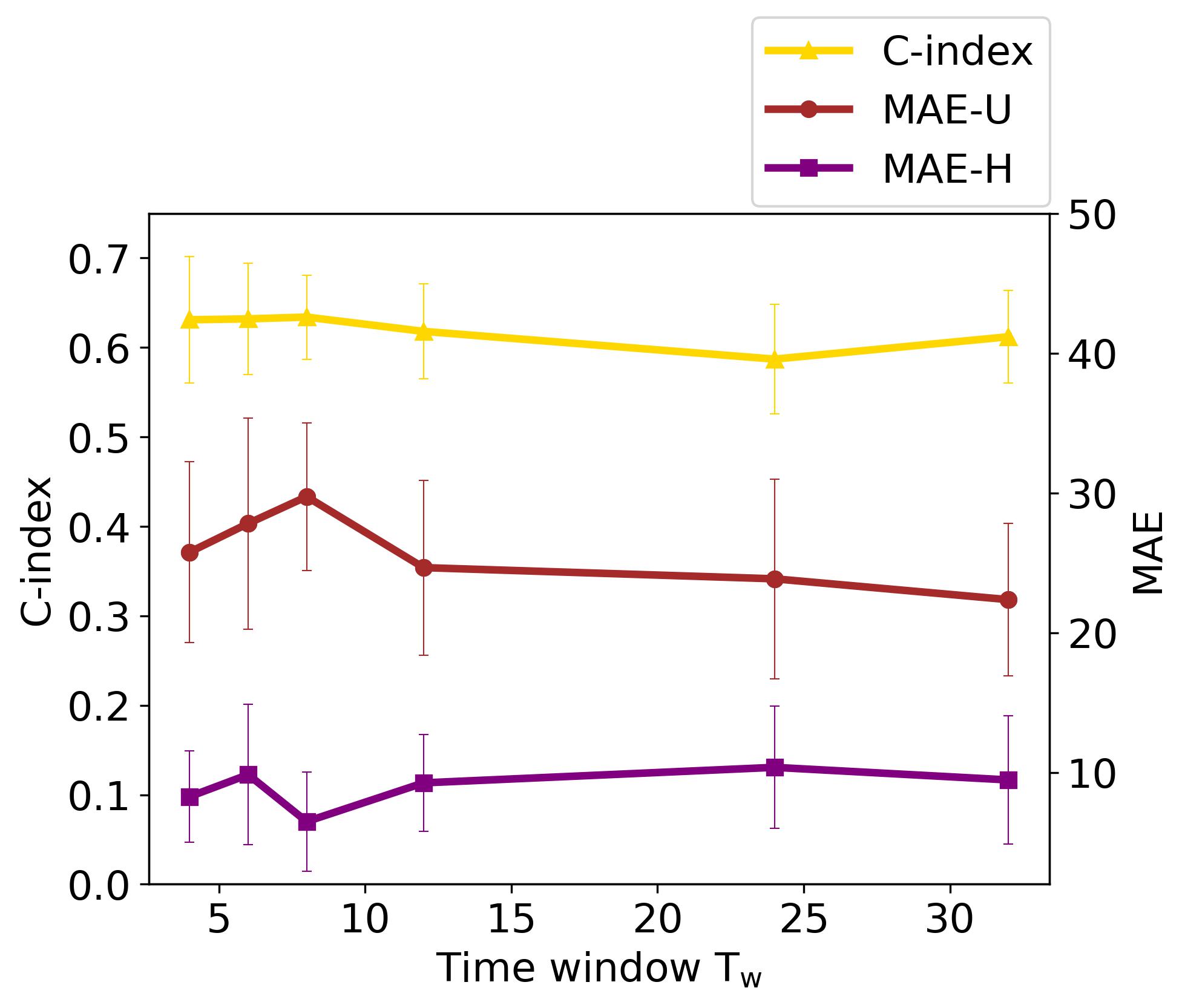}
}
\subfigure[$\lambda_m$ selection]{
\label{Lm_ablation}
\includegraphics[width=.225\textwidth]{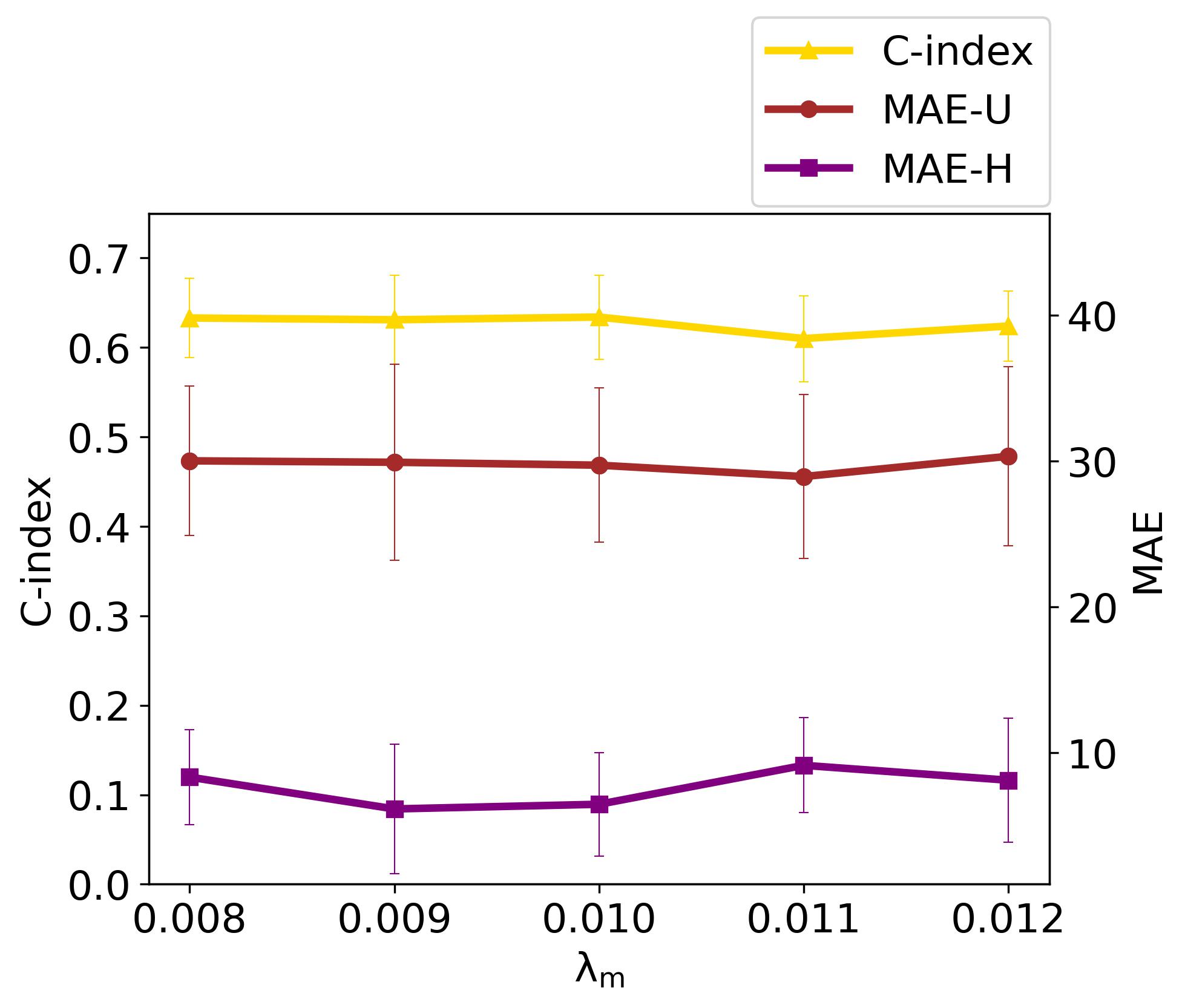}
}
\subfigure[$\lambda_v$ selection]{
\label{Lv_ablation}
\includegraphics[width=.225\textwidth]{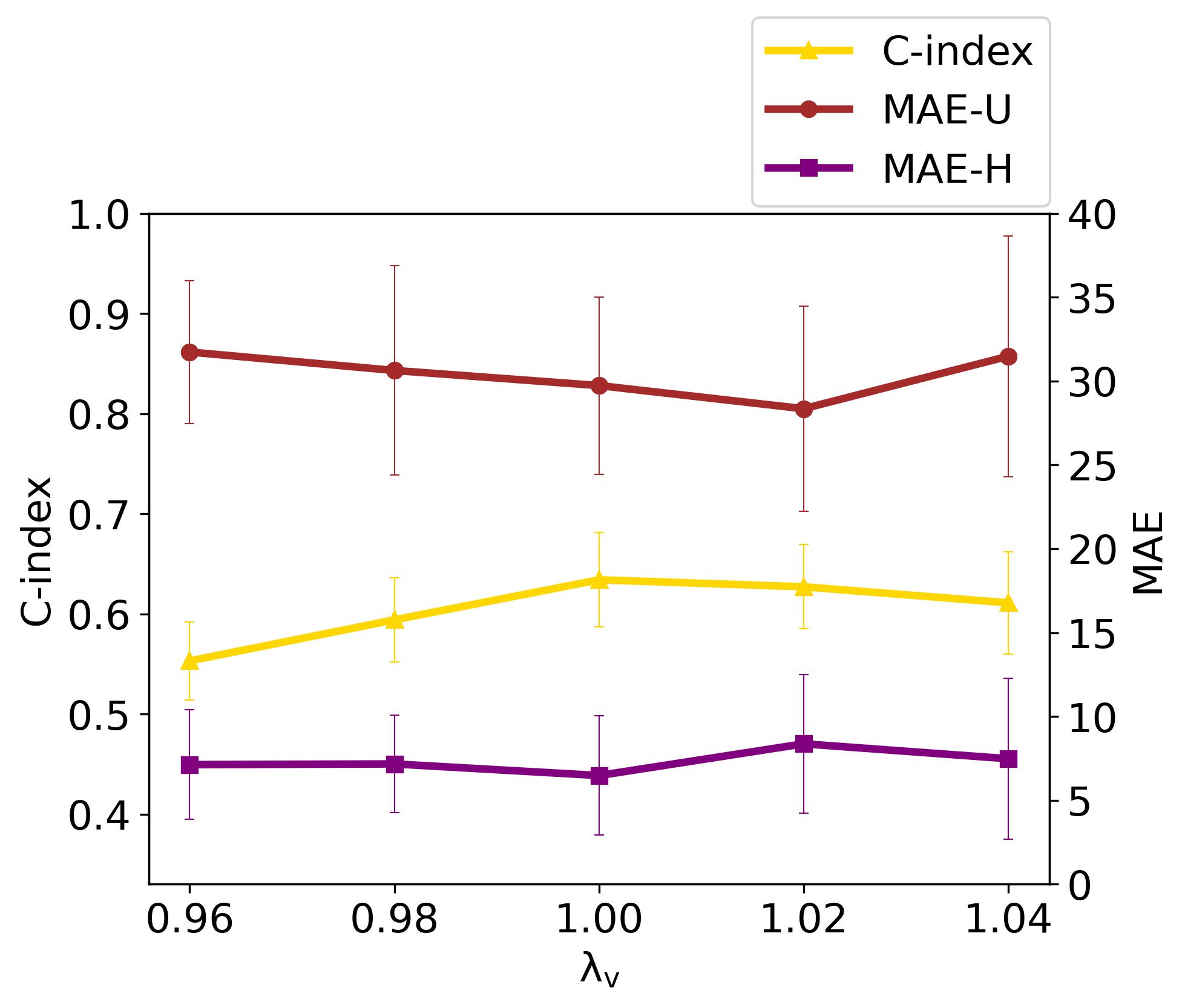}
}
\subfigure[Averaged PDF shape]{
\label{averaged_pdf}
\includegraphics[width=.23\textwidth]{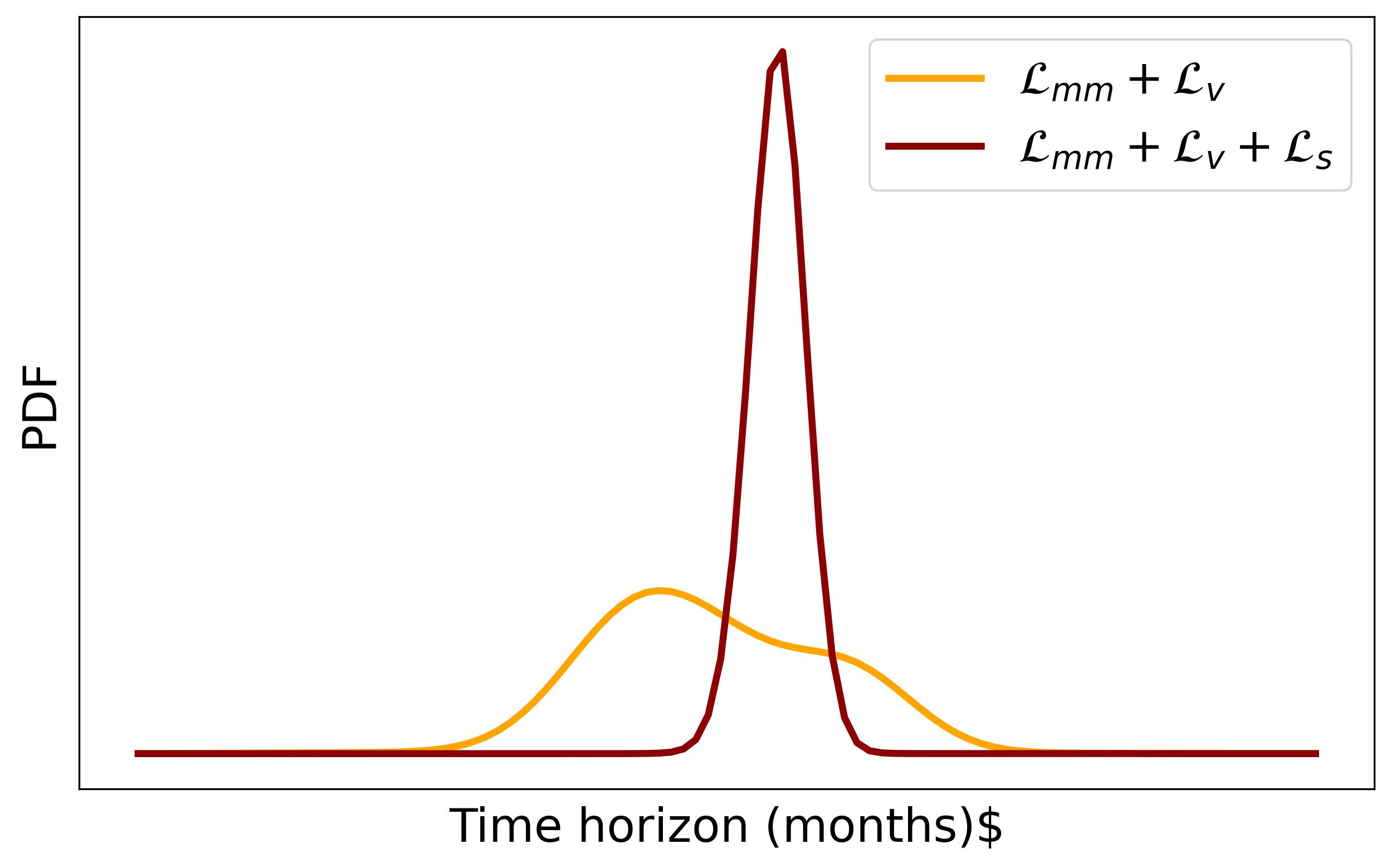}
}
\subfigure[PDF shape]{
\label{pdf_2}
\includegraphics[width=.27\textwidth]{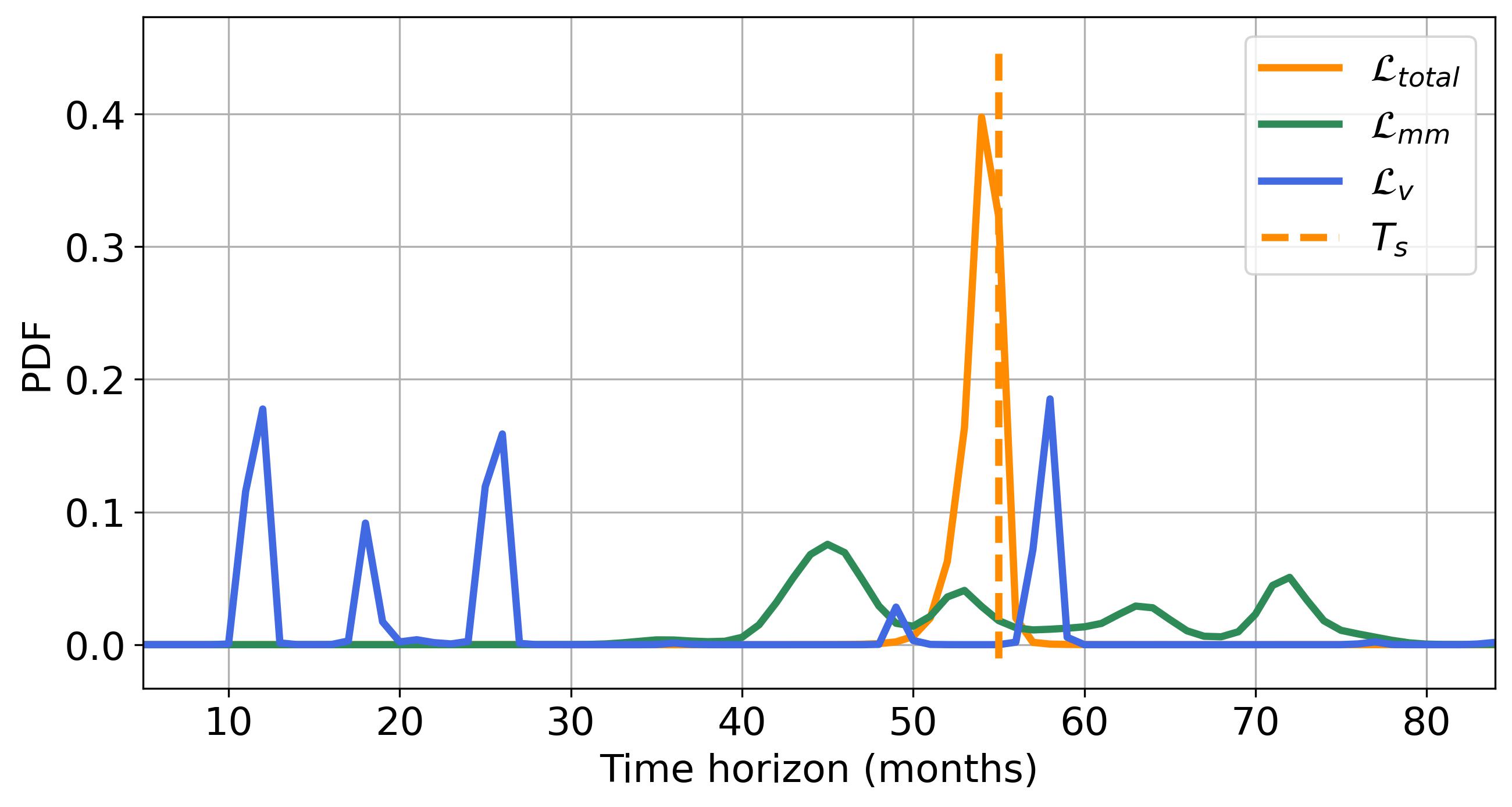}
}
\subfigure[PDF sensitivity]{
\label{pdf_3}
\includegraphics[width=.27\textwidth]{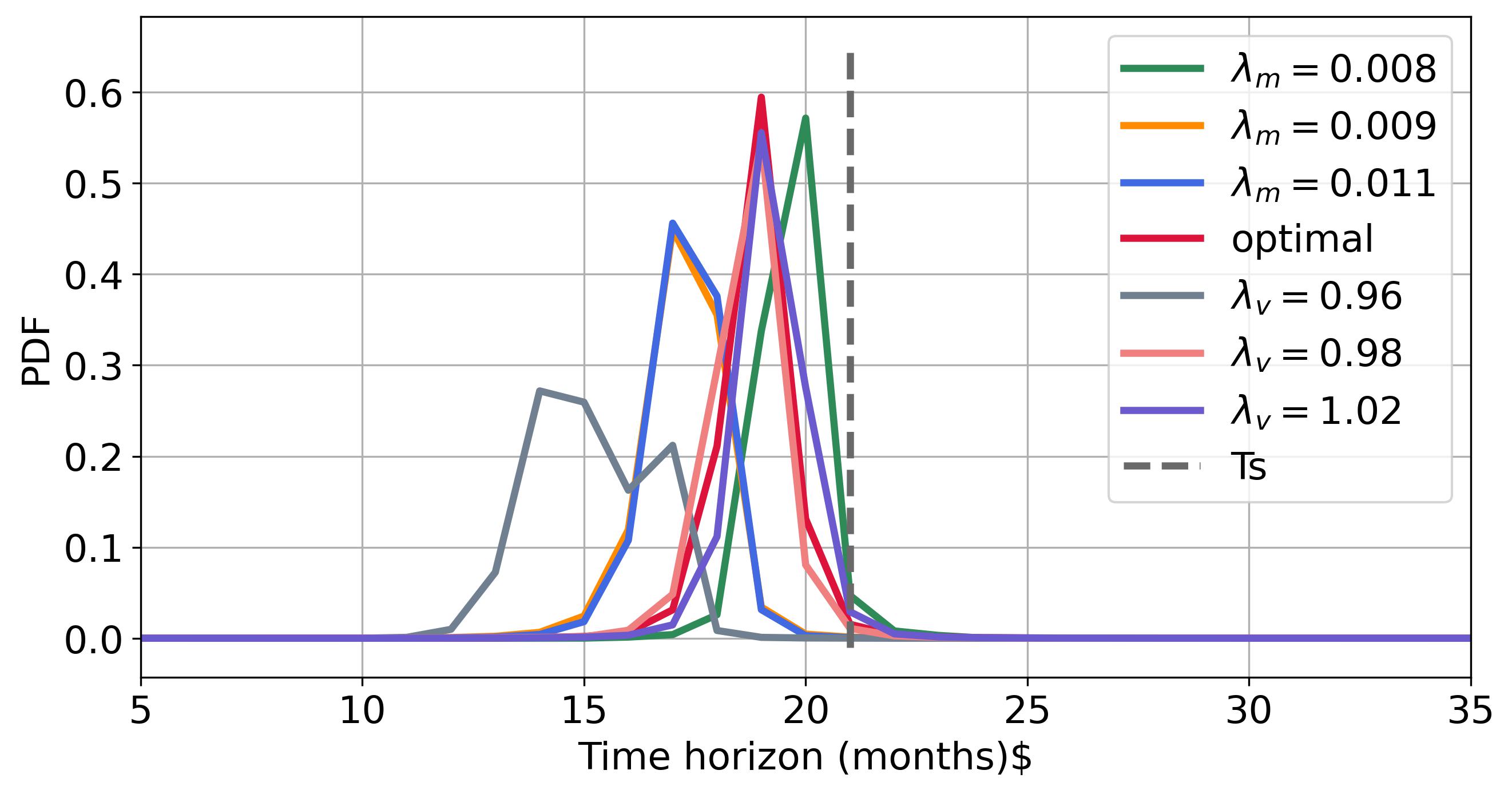}
}
\caption{Comparison results of the ablation study and the effectiveness analysis. All experiments are from UniSurv-d{\tiny 2} setting on MSReactor dataset}
\label{ablation}
\end{figure*}

\subsection{Ablation Study}
We further conduct an ablation study of losses on MSReactor using UniSurv-d{\tiny 2}, to demonstrate the contribution of each loss. In Fig. \ref{loss_ablation}, it is evident that an incomplete loss combination sometimes can lead to lower MAE-U or MAE-H, however, this often results in a situation of local optimization, which is reflected in the shape of the PDF. In Fig. \ref{pdf_2}, we compare the PDFs under selected scenarios: $\mathcal{L}_{mm}$ only, $\mathcal{L}_{v}$ only and $\mathcal{L}_{total}$. It is discernible that relying solely on $\mathcal{L}_{mm}$, due to the absence of $\mathcal{L}_{v}$ constraints, tends to produce a probability distribution biased towards uniformity around the ground truth. On the other hand, training solely with $\mathcal{L}_{v}$ generates irregular PDFs and fails to acquire meaningful information. This observation elucidates why these scenarios do not yield the optimal C-index.
Besides, the inclusion of $\mathcal{L}_{d}$ can further enhance the performance by mitigating the occurrence of discordant pairs.
In addition, incorporating $\mathcal{L}_{s}$ results in faster convergence and significant performance improvement, particularly evident in C-index. However, $\mathcal{L}_{s}$ also  leads to a rapid concentration of all probability distributions near the event time, which can result in overly concentrated PDFs and potential calibration issues. As shown in Fig. \ref{averaged_pdf}, we compared the averaged PDF shape with and without $\mathcal{L}_{s}$ in the combinations presented in Fig. \ref{loss_ablation}. The averaged shape is calculated by aligning all PDF peaks using the Dynamic Time Warp \citep{giorgino2009computing} technique and then averaging them in normalized horizon. It is apparent that the absence of $\mathcal{L}_{s}$ yields a multimodal, smoother, and more realistic PDF. Hence, the selection of $\mathcal{L}_{s}$ involves a trade-off between the ranking and calibration ability.

\subsection{Effectiveness Analysis}
\subsubsection{Sensitivity of Time Window $T_w$}
Fig. \ref{Tw_ablation} shows the effect of $T_w$. We can observe that when $T_w=8$, the model can achieve the highest C-index and lowest MAE-H, which is associated with the progression rate of MS. However, during the same period, MAE-U demonstrates its poorest performance. It is also apparent that the fluctuations in MAE-U and MAE-H exhibit a contrasting pattern. This disparity can be attributed to the disparate distributions of censoring and uncensoring within the MSReactor as in Fig. \ref{all_auc}. Meanwhile, this underscores that there exists potential for enhancing the robustness of UniSurv.

\subsubsection{Sensitivity of Loss Weights $\lambda_{m}$ And $\lambda_{v}$}
As the number of losses increases, finding the optimal weight combination indeed becomes challenging, but grid search can take care of this. The four losses do not need to be standardized to a similar magnitude. The unique characteristics of different datasets can lead to distinct optimal weights for losses. We assessed the sensitivities of $\lambda_{m}$ and $\lambda_{v}$ particularly on MSReactor in Fig. \ref{Lm_ablation} and Fig. \ref{Lv_ablation}, and some selected PDFs are shown in Fig. \ref{pdf_3}. The model exhibits robustness when small variations occur in $\lambda_{m}$ or $\lambda_{v}$, as the performance near their optimal values does not exhibit significant degradation. In some cases, two MAEs even perform better. This phenomenon is attributed to the opposing fluctuation trends exhibited by the MAEs, indicating a trade-off made by UniSurv during training. Notably, the C-index appears to be more sensitive to changes in $\lambda_{v}$ compared to variations in $\lambda_{m}$. As the variations in both weights increase, deviations in the PDF gradually emerge, with its peak drifting further away from the actual event time and assuming irregular shapes.

\subsubsection{Sensitivity of Larger and Noised Synthetic datasets}
To emphasize UniSurv's reliability for higher dimensionality datasets and its robustness to data noise, we have expanded the number of features for the existing synthetic datasets SYNTH-s and SYNTH-d without altering event or censoring settings, resulting in new SYNTH-s{\tiny k} and SYNTH-d{\tiny k} datasets, where k denotes the dimension of $\pmb{x}_{n}^{i}$ in Eq. \ref{synths_process} and Eq. \ref{synthd_process} is increased from $4$ to $4 \cdot 5^{k}$, and the dimension of $\pmb{x}_{v}^{i}$ in Eq. \ref{synthd_process} is increased from $20$ to $20 \cdot 5^{k}$. Additionally, we introduced noise $\pmb{\epsilon}^{i} \sim \epsilon_{0} \cdot \mathcal{N} (0, \mathbf{I})$ to $\pmb{x}_{n}^{i}$ and $\pmb{x}_{v}^{i}$ separately in all datasets. As the results shown in Tab. \ref{noised_synth}, UniSurv performs well on high-dimensional datasets and exhibits robustness to small levels of noise interference.

\begin{table*}[htbp]
\centering
\caption{Noise Sensitivity analysis on different sizes of synthetic datasets}
\resizebox{1\textwidth}{!}{%
\begin{tabular}{c|cccccc|cccccc}
\hline
\multirow{3}{*}{$\pmb{\epsilon_{0}}$} & \multicolumn{6}{c|}{\textbf{UniSurv-s}} & \multicolumn{6}{c}{\textbf{UniSurv-d{\tiny 2}}} \\ \cline{2-13} 
 & \multicolumn{2}{c|}{\textbf{SYNTH-s}} & \multicolumn{2}{c|}{\textbf{SYNTH-s$_{1}$} (k=1)} & \multicolumn{2}{c|}{\textbf{SYNTH-s$_{2}$}  (k=2)} & \multicolumn{2}{c|}{\textbf{SYNTH-d}} & \multicolumn{2}{c|}{\textbf{SYNTH-d$_{1}$}  (k=1)} & \multicolumn{2}{c}{\textbf{SYNTH-d$_{2}$}  (k=2)} \\ \cline{2-13} 
 & \textbf{C-index} & \multicolumn{1}{c|}{\textbf{MAE-H}} & \textbf{C-index} & \multicolumn{1}{c|}{\textbf{MAE-H}} & \textbf{C-index} & \textbf{MAE-H} & \textbf{C-index} & \multicolumn{1}{c|}{\textbf{MAE-H}} & \textbf{C-index} & \multicolumn{1}{c|}{\textbf{MAE-H}} & \textbf{C-index} & \textbf{MAE-H} \\ \hline
\textbf{$\pmb{0}$} & $0.731_{.008}$ & \multicolumn{1}{c|}{$1.55_{0.04}$} & $0.733_{.007}$ & \multicolumn{1}{c|}{$1.53_{0.05}$} & $0.732_{.008}$ & $1.54_{0.04}$ & $0.739_{.007}$ & \multicolumn{1}{c|}{$3.52_{0.08}$} & $0.740_{.008}$ & \multicolumn{1}{c|}{$3.51_{0.08}$} & $0.738_{.009}$ & $3.53_{0.09}$ \\ \hline
\textbf{$\pmb{0.1}$} & $0.731_{.008}$ & \multicolumn{1}{c|}{$1.56_{0.04}$} & $0.733_{.008}$ & \multicolumn{1}{c|}{$1.53_{0.05}$} & $0.732_{.009}$ & $1.55_{0.04}$ & $0.739_{.007}$ & \multicolumn{1}{c|}{$3.53_{0.08}$} & $0.740_{.008}$ & \multicolumn{1}{c|}{$3.51_{0.09}$} & $0.737_{.010}$ & $3.53_{0.09}$ \\ \hline
$\pmb{0.3}$ & $0.729_{.010}$ & \multicolumn{1}{c|}{$1.56_{0.05}$} & $0.731_{.009}$ & \multicolumn{1}{c|}{$1.54_{0.06}$} & $0.729_{.009}$ & $1.57_{0.05}$ & $0.738_{.006}$ & \multicolumn{1}{c|}{$3.54_{0.09}$} & $0.738_{.007}$ & \multicolumn{1}{c|}{$3.53_{0.08}$} & $0.735_{.009}$ & $3.56_{0.07}$ \\ \hline
\end{tabular}%
}
\label{noised_synth}
\end{table*}

\section{Conclusion And Discussion}\label{sec5}

In this paper, we propose a non-parametric discrete survival model named UniSurv. Departing from the existing models of utilizing RNN for processing longitudinal data, we employ a Transformer for adeptly handling dynamic analysis. In particular, our survival framework firstly integrates imputation for handling missing data issue, then incorporates different embedding branches for time-varying and time-invariant features extraction. The Transformer encoder takes merged features as input and outputs the individual PDF. We also demonstrated how to process image-like data using variations of modules and how to select a time window based on the progression speed of the disease to share information. This is particularly beneficial in the field of medicine, as obtaining regular time-series medical images in the real world is challenging.

Furthermore, our novel Margin-Mean-Variance loss effectively produces smooth PDF in a unimodal manner, demonstrating clear superiority over other discrete models. Importantly, the proposed loss can be seamlessly embedded into various discrete survival models. Moreover, it significantly enhances prediction accuracy, particularly for patients with extended censoring times. Applying poorly performing models in such scenarios could evidently disrupt physician's judgments and place unnecessary burdens on both society and healthcare institutions. This constitutes a valuable contribution. Although our current PDF may appear overly concentrated around event times, akin to many models relying on strong probability assumptions, resulting in unconventional survival curves, we intend to further modify the $\mathcal{L}_{s}$ and $\mathcal{L}_{v}$ to relax certain constraints in the future. This adjustment aims to yield a more elegant PDF, characterized by a smoother and less abrupt distribution while maintaining overall performance. Meanwhile, adapting UniSurv to accommodate multiple censoring scenarios, such as left truncation and interval-censored data, presents an interesting direction for future research. Additionally, expanding the scope to include a post-processing statistic for interpreting risk predictions in both static and dynamic analyses of disease progression is necessary. For example, individual explanations of predicted probabilities can be achieved through the generation of SHapley Additive exPlanations (SHAP) \citep{pieszko2023time,krzyzinski2023survshap}. This approach is expected to result in more effective health care.

\backmatter

\bmhead{Acknowledgements}

X.Z. receives support from the Australian Government Research Training Program (RTP) Scholarship.

\section*{Declarations}
\textbf{Availability of data and materials} We are restricted from making MSReactor data available to the public for the moment. All the other data are publicly available. \newline
\textbf{Competing interests} Not applicable. \newline
\textbf{Ethics approval} Not applicable. \newline
\textbf{Consent for participation} Not applicable. \newline
\textbf{Consent for publication} Not applicable. \newline

\begin{appendices}

\section{MSReactor}\label{secA2}
\subsection{Missing Details}
For each patient, temporal tests were done once around roughly every half-year. The time interval between two adjacent tests range from $0$ to $44$ months with mean of $4.68$. The number of yearly follow-ups was from $1$ to $6$ with mean of $2.20$ tests per patients. 

\subsection{Min-Max Values Selection of Reaction Tensor Representation}
In our proposed innovative reaction tensor representation, as illustrated in Fig. \ref{data_rep}, the chosen minimum and maximum values are not directly derived from the original tabular data for MSReactor dataset, but rather determined by threshold selection. Specifically, we sort individual patient's data for a particular task $c$, and the $2\%$ and $98\%$ percentiles of the sorted values are taken as the minimum and maximum values, denoted as $\alpha_{min\_thr, c}$ and $\alpha_{max\_thr, c}$, respectively. The outliers are automatically set as minimum or maximum values. We use the same strategy in SYNTH-d dataset.

The rationale for adopting this approach stems from the inherent instability of recorded reaction times in such tests, attributed at times to individual patient idiosyncrasies or extraneous environmental interference, rendering these reaction times as outliers in our analysis. Instances of a patient expending time to accommodate their sitting posture, being diverted by ambient noise distractions, or experiencing rapid inadvertent touchscreen interactions, exemplify scenarios capable of inducing aberrations in reaction time.

\section{Hyperparameter Information}\label{secA4}
Tab. \ref{hyperparameter} shows the hyperparameter spaces and their optimal choices we used in UniSurv-s for SUPPORT, METABRIC and SYNTH-s dataset, and in UniSurv-d{\tiny2} for SYNTH-d, MSReactor. Clearly, we shrink the architecture of Transformer encoder part, since the survival datasets are much smaller than the standard natural language processing (NLP) or computer vision (CV) datasets, and the number of features is also small. Besides, the selection of $T_{max}$ is based on the maximum survival/censoring time in the corresponding dataset, and the value of $T_{w}$ is directly chosen from the factors of $T_{max}+1$ for the sake of convenience during training dynamic datasets.
\begin{table*}[htbp]
\centering
\caption{The hyperparameter spaces for five datasets. We bold the \textbf{optimal choice}}
\resizebox{1\textwidth}{!}{%
\begin{tabular}{c|c|c|c|c|c}
\hline
\textbf{Hyperparameter} & \textbf{SUPPORT} & \textbf{METABRIC} & \textbf{SYNTH-s} & \textbf{MSReactor} & \textbf{SYNTH-d} \\ \hline
$T_{max}$ & \pmb{80} & \pmb{400} & \pmb{200} & \pmb{95} & \pmb{199} \\ \hline
$T_w$ & - & - & - & \{4, 6, \pmb{8}, 12, 24, 32\} & \{4, 5, \pmb{8}, 25, 40\} \\ \hline
$\lambda_{m}$ & \{\pmb{0.01}, 0.1, 1, 10\} & \{\pmb{0.01}, 0.1, 1, 10\} & \{0.01, 0.1, \pmb{1}, 10\} & \{\pmb{0.01}, 0.1, 1, 10\} & \{0.01, \pmb{0.1}, 1, 10\} \\ \hline
$\lambda_{v}$ & \{\pmb{0.001}, 0.01, 0.1, 1\} & \{\pmb{0.001}, 0.01, 0.1, 1\} & \{0.001, \pmb{0.01}, 0.1, 1\} & \{0.001, 0.01, 0.1, \pmb{1}\} & \{0.001, 0.01, \pmb{0.1}, 1\} \\ \hline
$\lambda_{d}$ & \{0, \pmb{1}\} & \{0, \pmb{1}\} & \{0, \pmb{1}\} & \{0, \pmb{1}\} & \{0, \pmb{1}\} \\ \hline
Epochs & \pmb{400} & \pmb{200} & \pmb{200} & \pmb{200} & \pmb{200} \\ \hline
Batch size & \pmb{16} & \pmb{16} & \{4, 8, \pmb{16}, 32\} & \{4, 8, \pmb{16}, 32\} & \{4, 8, \pmb{16}, 32\} \\ \hline
Dropout rate & \pmb{0.1} & \pmb{0.1} & \{0.0, \pmb{0.1}, 0.3\} & \{0.0, \pmb{0.1}, 0.3\} & \{0.0, \pmb{0.1}, 0.3\} \\ \hline
Number of heads & \pmb{4} & \pmb{4} & \{1, 2, \pmb{4}, 8\} & \{1, 2, 4, \pmb{8}\} & \{1, 2, \pmb{4}, 8\} \\ \hline
Embedding dimension & \pmb{512} & \pmb{512} & \{\pmb{256}, 512\} & \{256, \pmb{512}\} & \{\pmb{256}, 512\} \\ \hline
Number of attention layers & \pmb{4} & \pmb{4} & \{1, 2, 3, \pmb{4}\} & \{1, 2, 3, \pmb{4}\} & \{1, 2, 3, \pmb{4}\} \\ \hline
Adam optimizer with fixed learning rate & \pmb{1e-4} & \pmb{1e-4} & \{\pmb{1e-4}, 1e-3\} & \{\pmb{1e-4}, 1e-3\} & \{\pmb{1e-4}, 1e-3\} \\ \hline
\end{tabular}
}
\label{hyperparameter}
\end{table*}

\end{appendices}


\bibliography{sn-bibliography}

\begin{thebibliography}{37}
\providecommand{\natexlab}[1]{#1}
\providecommand{\url}[1]{{#1}}
\providecommand{\urlprefix}{URL }
\providecommand{\doi}[1]{\url{https://doi.org/#1}}
\providecommand{\eprint}[2][]{\url{#2}}
 \bibcommenthead

\bibitem[{Collett(2023)}]{collett2023modelling}
Collett D (2023) Modelling survival data in medical research. CRC press

\bibitem[{Cox(1972)}]{cox1972regression}
Cox DR (1972) Regression models and life-tables. Journal of the Royal Statistical Society: Series B (Methodological) 34(2):187--202

\bibitem[{Curtis et~al(2012)Curtis, Shah, Chin, Turashvili, Rueda, Dunning, Speed, Lynch, Samarajiwa, Yuan et~al}]{curtis2012genomic}
Curtis C, Shah SP, Chin SF, et~al (2012) The genomic and transcriptomic architecture of 2,000 breast tumours reveals novel subgroups. Nature 486(7403):346--352

\bibitem[{Davidson-Pilon(2019)}]{Davidson-Pilon2019}
Davidson-Pilon C (2019) lifelines: survival analysis in python. Journal of Open Source Software 4(40):1317. \doi{10.21105/joss.01317}, \urlprefix\url{https://doi.org/10.21105/joss.01317}

\bibitem[{Faraggi and Simon(1995)}]{faraggi1995neural}
Faraggi D, Simon R (1995) A neural network model for survival data. Statistics in medicine 14(1):73--82

\bibitem[{Foong et~al(2023)Foong, Bridge, Merlo, Gresle, Zhu, Buzzard, Butzkueven, and van~der Walt}]{foong2023smartphone}
Foong YC, Bridge F, Merlo D, et~al (2023) Smartphone monitoring of cognition in people with multiple sclerosis: A systematic review. Multiple Sclerosis and Related Disorders p 104674

\bibitem[{Giorgino(2009)}]{giorgino2009computing}
Giorgino T (2009) Computing and visualizing dynamic time warping alignments in r: the dtw package. Journal of statistical Software 31:1--24

\bibitem[{Haider et~al(2020)Haider, Hoehn, Davis, and Greiner}]{haider2020effective}
Haider H, Hoehn B, Davis S, et~al (2020) Effective ways to build and evaluate individual survival distributions. The Journal of Machine Learning Research 21(1):3289--3351

\bibitem[{Hu et~al(2021)Hu, Fridgeirsson, van Wingen, and Welling}]{hu2021transformer}
Hu S, Fridgeirsson E, van Wingen G, et~al (2021) Transformer-based deep survival analysis. In: Survival Prediction-Algorithms, Challenges and Applications, PMLR, pp 132--148

\bibitem[{Hunter et~al(2021)Hunter, Aburashed, Alroughani, Chan, Dive, Eichau, Kantor, Kim, Lycke, Macdonell et~al}]{hunter2021confirmed}
Hunter SF, Aburashed RA, Alroughani R, et~al (2021) Confirmed 6-month disability improvement and worsening correlate with long-term disability outcomes in alemtuzumab-treated patients with multiple sclerosis: Post hoc analysis of the care-ms studies. Neurology and therapy 10(2):803--818

\bibitem[{Ishwaran et~al(2008)Ishwaran, Kogalur, Blackstone, and Lauer}]{ishwaran2008random}
Ishwaran H, Kogalur UB, Blackstone EH, et~al (2008) Random survival forests. The annals of applied statistics 2(3):841--860

\bibitem[{Kaplan and Meier(1958)}]{kaplan1958nonparametric}
Kaplan EL, Meier P (1958) Nonparametric estimation from incomplete observations. Journal of the American statistical association 53(282):457--481

\bibitem[{Katzman et~al(2018)Katzman, Shaham, Cloninger, Bates, Jiang, and Kluger}]{katzman2018deepsurv}
Katzman JL, Shaham U, Cloninger A, et~al (2018) Deepsurv: personalized treatment recommender system using a cox proportional hazards deep neural network. BMC medical research methodology 18(1):1--12

\bibitem[{Knaus et~al(1995)Knaus, Harrell, Lynn, Goldman, Phillips, Connors, Dawson, Fulkerson, Califf, Desbiens et~al}]{knaus1995support}
Knaus WA, Harrell FE, Lynn J, et~al (1995) The support prognostic model: Objective estimates of survival for seriously ill hospitalized adults. Annals of internal medicine 122(3):191--203

\bibitem[{Krzyzi{\'n}ski et~al(2023)Krzyzi{\'n}ski, Spytek, Baniecki, and Biecek}]{krzyzinski2023survshap}
Krzyzi{\'n}ski M, Spytek M, Baniecki H, et~al (2023) Survshap (t): time-dependent explanations of machine learning survival models. Knowledge-Based Systems 262:110234

\bibitem[{Lambert and Chevret(2016)}]{lambert2016summary}
Lambert J, Chevret S (2016) Summary measure of discrimination in survival models based on cumulative/dynamic time-dependent roc curves. Statistical methods in medical research 25(5):2088--2102

\bibitem[{Lee et~al(2018)Lee, Zame, Yoon, and Van Der~Schaar}]{lee2018deephit}
Lee C, Zame W, Yoon J, et~al (2018) Deephit: A deep learning approach to survival analysis with competing risks. In: Proceedings of the AAAI conference on artificial intelligence

\bibitem[{Lee et~al(2019)Lee, Yoon, and Van Der~Schaar}]{lee2019dynamic}
Lee C, Yoon J, Van Der~Schaar M (2019) Dynamic-deephit: A deep learning approach for dynamic survival analysis with competing risks based on longitudinal data. IEEE Transactions on Biomedical Engineering 67(1):122--133

\bibitem[{Lee and Whitmore(2006)}]{lee2006threshold}
Lee MLT, Whitmore G (2006) Threshold regression for survival analysis: Modeling event times by a stochastic process reaching a boundary. Statist Sci 21(1):501--513

\bibitem[{Leung et~al(1997)Leung, Elashoff, and Afifi}]{leung1997censoring}
Leung KM, Elashoff RM, Afifi AA (1997) Censoring issues in survival analysis. Annual review of public health 18(1):83--104

\bibitem[{Luck et~al(2017)Luck, Sylvain, Cardinal, Lodi, and Bengio}]{luck2017deep}
Luck M, Sylvain T, Cardinal H, et~al (2017) Deep learning for patient-specific kidney graft survival analysis. arXiv preprint arXiv:170510245

\bibitem[{Merlo et~al(2021)Merlo, Stankovich, Bai, Kalincik, Zhu, Gresle, Lechner-Scott, Kilpatrick, Barnett, Taylor et~al}]{merlo2021association}
Merlo D, Stankovich J, Bai C, et~al (2021) Association between cognitive trajectories and disability progression in patients with relapsing-remitting multiple sclerosis. Neurology 97(20):e2020--e2031

\bibitem[{Nagpal et~al(2021{\natexlab{a}})Nagpal, Jeanselme, and Dubrawski}]{nagpal2021deep2}
Nagpal C, Jeanselme V, Dubrawski A (2021{\natexlab{a}}) Deep parametric time-to-event regression with time-varying covariates. In: Survival Prediction-Algorithms, Challenges and Applications, PMLR, pp 184--193

\bibitem[{Nagpal et~al(2021{\natexlab{b}})Nagpal, Li, and Dubrawski}]{nagpal2021deep}
Nagpal C, Li X, Dubrawski A (2021{\natexlab{b}}) Deep survival machines: Fully parametric survival regression and representation learning for censored data with competing risks. IEEE Journal of Biomedical and Health Informatics 25(8):3163--3175

\bibitem[{Nagpal et~al(2021{\natexlab{c}})Nagpal, Yadlowsky, Rostamzadeh, and Heller}]{nagpal2021deep3}
Nagpal C, Yadlowsky S, Rostamzadeh N, et~al (2021{\natexlab{c}}) Deep cox mixtures for survival regression. In: Machine Learning for Healthcare Conference, PMLR, pp 674--708

\bibitem[{Pan et~al(2018)Pan, Han, Shan, and Chen}]{pan2018mean}
Pan H, Han H, Shan S, et~al (2018) Mean-variance loss for deep age estimation from a face. In: Proceedings of the IEEE conference on computer vision and pattern recognition, pp 5285--5294

\bibitem[{Pham et~al(2021)Pham, Harris, Varosanec, Morgan, Kosa, and Bielekova}]{pham2021smartphone}
Pham L, Harris T, Varosanec M, et~al (2021) Smartphone-based symbol-digit modalities test reliably captures brain damage in multiple sclerosis. NPJ digital medicine 4(1):36

\bibitem[{Pieszko et~al(2023)Pieszko, Shanbhag, Singh, Hauser, Miller, Liang, Motwani, Kwieci{\'n}ski, Sharir, Einstein et~al}]{pieszko2023time}
Pieszko K, Shanbhag AD, Singh A, et~al (2023) Time and event-specific deep learning for personalized risk assessment after cardiac perfusion imaging. npj Digital Medicine 6(1):78

\bibitem[{P{\"o}lsterl(2020)}]{sksurv}
P{\"o}lsterl S (2020) scikit-survival: A library for time-to-event analysis built on top of scikit-learn. Journal of Machine Learning Research 21(212):1--6. \urlprefix\url{http://jmlr.org/papers/v21/20-729.html}

\bibitem[{Pourjafari et~al(2022)Pourjafari, Ziaei, Rezaei, Sameizadeh, Shafiee, Alavinia, Abolghasemian, and Sajadi}]{pourjafari2022survival}
Pourjafari E, Ziaei N, Rezaei MR, et~al (2022) Survival seq2seq: A survival model based on sequence to sequence architecture. In: Machine Learning for Healthcare Conference, PMLR, pp 79--100

\bibitem[{Rindt et~al(2022)Rindt, Hu, Steinsaltz, and Sejdinovic}]{rindt2022survival}
Rindt D, Hu R, Steinsaltz D, et~al (2022) Survival regression with proper scoring rules and monotonic neural networks. In: International Conference on Artificial Intelligence and Statistics, PMLR, pp 1190--1205

\bibitem[{Singer and Willett(1991)}]{singer1991modeling}
Singer JD, Willett JB (1991) Modeling the days of our lives: using survival analysis when designing and analyzing longitudinal studies of duration and the timing of events. psychological Bulletin 110(2):268

\bibitem[{Uno et~al(2011)Uno, Cai, Pencina, D'Agostino, and Wei}]{uno2011c}
Uno H, Cai T, Pencina MJ, et~al (2011) On the c-statistics for evaluating overall adequacy of risk prediction procedures with censored survival data. Statistics in medicine 30(10):1105--1117

\bibitem[{Vaswani et~al(2017)Vaswani, Shazeer, Parmar, Uszkoreit, Jones, Gomez, Kaiser, and Polosukhin}]{vaswani2017attention}
Vaswani A, Shazeer N, Parmar N, et~al (2017) Attention is all you need. Advances in neural information processing systems 30

\bibitem[{Vinzamuri and Reddy(2013)}]{vinzamuri2013cox}
Vinzamuri B, Reddy CK (2013) Cox regression with correlation based regularization for electronic health records. In: 2013 IEEE 13th International Conference on Data Mining, IEEE, pp 757--766

\bibitem[{Wang and Sun(2022)}]{wang2022survtrace}
Wang Z, Sun J (2022) Survtrace: Transformers for survival analysis with competing events. In: Proceedings of the 13th ACM International Conference on Bioinformatics, Computational Biology and Health Informatics, pp 1--9

\bibitem[{Whitehouse et~al(2019)Whitehouse, Fisk, Bernstein, Berrigan, Bolton, Graff, Hitchon, Marriott, Peschken, Sareen et~al}]{whitehouse2019comorbid}
Whitehouse CE, Fisk JD, Bernstein CN, et~al (2019) Comorbid anxiety, depression, and cognition in ms and other immune-mediated disorders. Neurology 92(5):e406--e417

\end{thebibliography}

\end{document}